%% file: main.tex
\documentclass[runningheads]{llncs}

\usepackage{eccv}

\usepackage{eccvabbrv}

\usepackage{graphicx}
\usepackage{booktabs}

\usepackage[accsupp]{axessibility}  %

\usepackage{hyperref}

\usepackage{orcidlink}

\usepackage{amsmath}
\DeclareMathOperator*{\argmin}{arg\,min}

\usepackage{amsfonts}
\usepackage{bm}
\usepackage{graphicx}
\usepackage[T1]{fontenc}

\usepackage{csquotes}
\usepackage{multirow}
\usepackage{caption}

\usepackage{subcaption}
\usepackage{makecell}

\usepackage{algorithm}
\usepackage{algpseudocode}

\usepackage{stackengine}
\newcommand\stackequal[2]{%
  \mathrel{\stackunder{\stackon{$=$}{$\scriptstyle#1$}}{%
  $\scriptstyle#2$}}}

\begin{document}

\title{Bayesian Detector Combination for Object Detection with Crowdsourced Annotations}

\titlerunning{Bayesian Detector Combination}

\author{Zhi Qin Tan \inst{1}\orcidlink{0000-0002-5521-6808} \and
Olga Isupova\inst{2}\orcidlink{0000-0002-8506-6871} \and
Gustavo Carneiro\inst{1}\orcidlink{0000-0002-5571-6220} \and
Xiatian Zhu\inst{1}\orcidlink{0000-0002-9284-2955} \and
Yunpeng Li\inst{1}\orcidlink{0000-0003-4798-541X}
}

\authorrunning{Z.Q. Tan et al.}

\institute{
University of Surrey, Surrey, UK
\email{\{z.tan,g.carneiro,xiatian.zhu,yunpeng.li\}@surrey.ac.uk} \and
University of Oxford, Oxford, UK \\
\email{olga.isupova@ouce.ox.ac.uk} 
}

\maketitle

\algnewcommand{\LineComment}[1]{\State \(\triangleright\) #1}
\algrenewcommand\algorithmicensure{\textbf{Output:}}
\newcommand*\diff{\mathop{}\!\mathrm{d}}
\newcommand{\mathbbm}[1]{\text{\usefont{U}{bbm}{m}{n}#1}}

\input{sec/0_abstract}    
\input{sec/1_intro}

\input{sec/2_lit_review}
\input{sec/3_methodology}

\input{sec/4_5_exp_and_discuss}

\input{sec/6_conclusion}

\paragraph{\bf Acknowledgements.} Z.Q. T. acknowledges the support of an Industrial Cooperative Awards in Science and Engineering (ICASE) studentship from the Engineering and Physical Sciences Research Council (EPSRC) for this work. G. C. acknowledges the support of the Engineering and Physical Sciences Research Council (EPSRC) through grant EP/Y018036/1 and the Australian Research Council (ARC) through grant FT190100525. The authors would like to thank the Satellite Applications Catapult for the provision of the disaster response dataset.

\bibliographystyle{splncs04}
\bibliography{main}

\clearpage
\begin{center}
    {\Large \bfseries\boldmath
      \pretolerance=10000
  Supplementary Material \par}\vskip .8cm
\end{center}
\appendix
\input{sec/X_suppl}

\end{document}

%% file: sec/0_abstract.tex
\begin{abstract}

Acquiring fine-grained object detection annotations in unconstrained images is time-consuming, expensive, and prone to noise, especially in crowdsourcing scenarios. Most prior object detection methods assume accurate annotations; A few recent works have studied object detection with noisy crowdsourced annotations, with evaluation on distinct synthetic crowdsourced datasets of varying setups under artificial assumptions. %
To address these algorithmic limitations and evaluation inconsistency, we first propose a novel Bayesian Detector Combination (BDC) framework to more effectively train object detectors with noisy crowdsourced annotations, with the unique ability of automatically inferring the annotators' label qualities.
Unlike previous approaches, BDC is model-agnostic, requires no prior knowledge of the annotators' skill level, and seamlessly integrates with existing object detection models. 
Due to the scarcity of real-world crowdsourced datasets, we introduce large synthetic datasets by simulating varying crowdsourcing scenarios. This allows consistent evaluation of different models at scale.
Extensive experiments on both real and synthetic crowdsourced datasets show that BDC outperforms existing state-of-the-art methods, demonstrating its superiority in leveraging crowdsourced data for object detection.
Our code and data are available at: \url{https://github.com/zhiqin1998/bdc}.

\keywords{Bayesian Detector Combination \and Object detection \and Crowdsourced annotations}
\end{abstract}

%% file: sec/1_intro.tex
\section{Introduction}
\label{sect:intro}

\begin{figure}[tb]
    \centering
    \subcaptionbox{ID: 70282}{\includegraphics[width=0.19\linewidth,trim={2cm 12.5cm 2cm 2.5cm},clip]{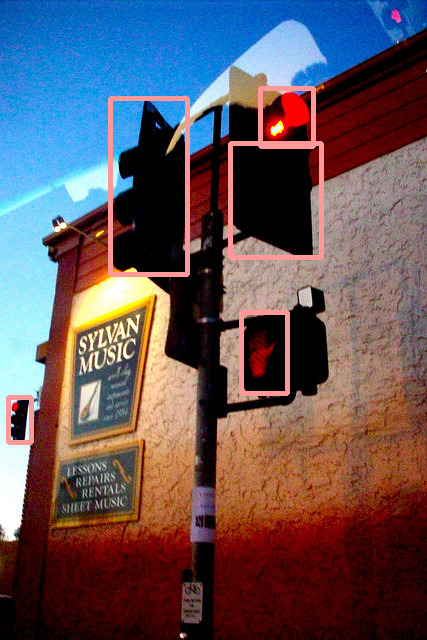}}
    \subcaptionbox{ID: 310072}{\includegraphics[width=0.19\linewidth,trim={0 2cm 5.5cm 0},clip]{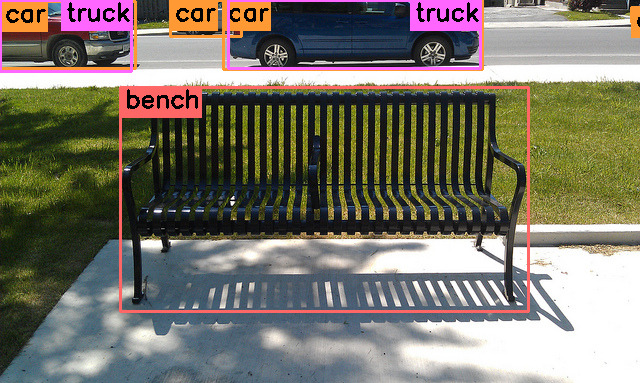}}
    \subcaptionbox{ID: 516471}{\includegraphics[width=0.19\linewidth,trim={2.5cm 1cm 2cm 3cm},clip]{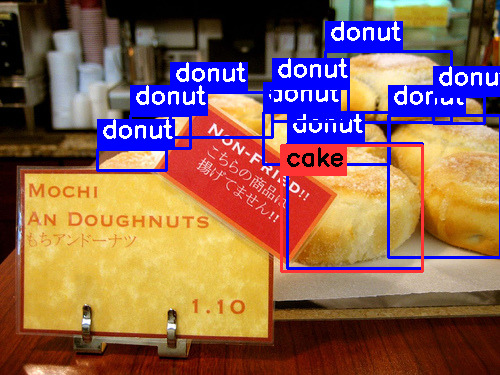}}
    \subcaptionbox{ID: 326247}{\includegraphics[width=0.19\linewidth,trim={3cm 2cm 0 1cm},clip]{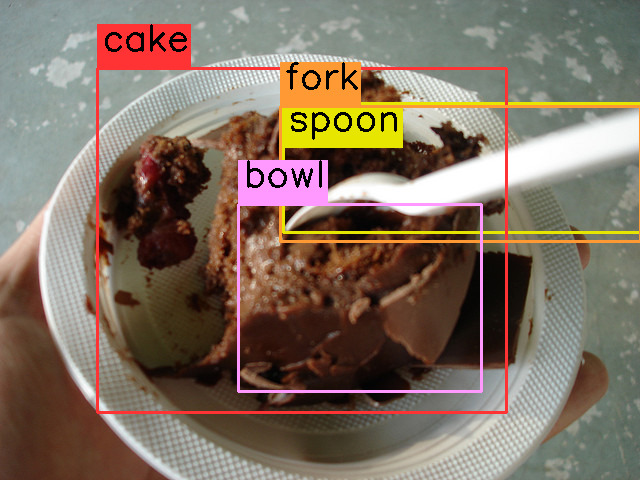}}
    \subcaptionbox{ID: 416149}{\includegraphics[width=0.19\linewidth,trim={0cm 9.5cm 0cm 1cm},clip]{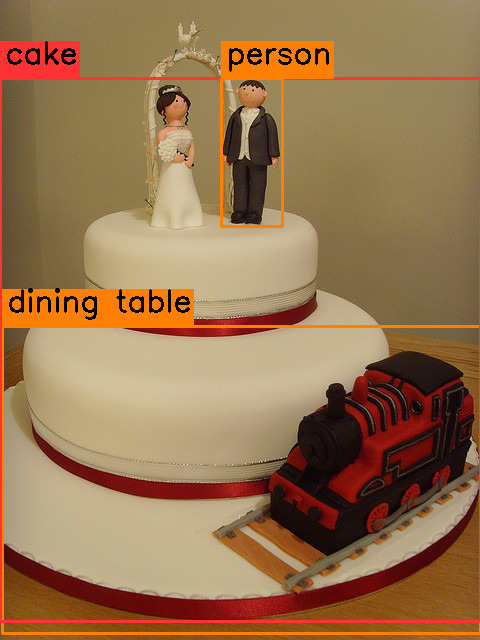}}

    \subcaptionbox{\label{fig:vindr_example1}}{\includegraphics[width=0.19\linewidth,trim={0.5cm 1.8cm 0cm 2.7cm},clip]{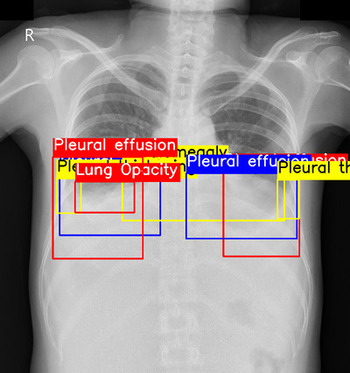}}
    \subcaptionbox{\label{fig:vindr_example2}}{\includegraphics[width=0.19\linewidth,trim={1cm 3.8cm 1cm 1cm},clip]{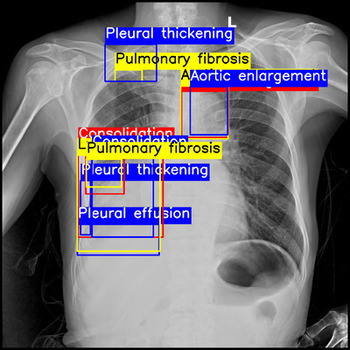}}
    \subcaptionbox{\label{fig:vindr_example3}}{\includegraphics[width=0.19\linewidth,trim={1cm 2.3cm 0cm 0.5cm},clip]{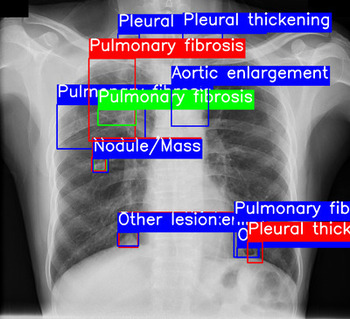}}
    \subcaptionbox{\label{fig:vindr_example4}}{\includegraphics[width=0.19\linewidth,trim={0cm 3.7cm 0cm 2.5cm},clip]{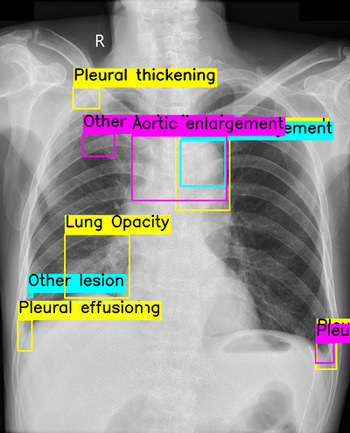}}
    \subcaptionbox{\label{fig:vindr_example5}}{\includegraphics[width=0.19\linewidth,trim={0cm 4.1cm 0cm 1cm},clip]{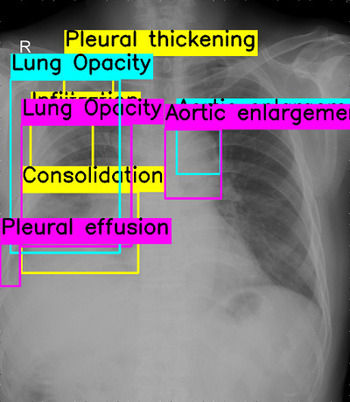}}

    \subcaptionbox{\label{fig:dis_example1}}{\includegraphics[width=0.19\linewidth,trim={0cm 0cm 9cm 13cm},clip]{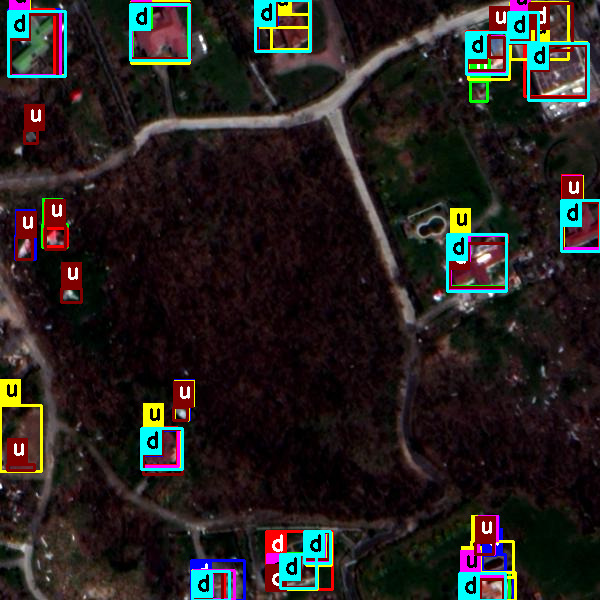}}
    \subcaptionbox{\label{fig:dis_example2}}{\includegraphics[width=0.19\linewidth,trim={6cm 0cm 4cm 13.7cm},clip]{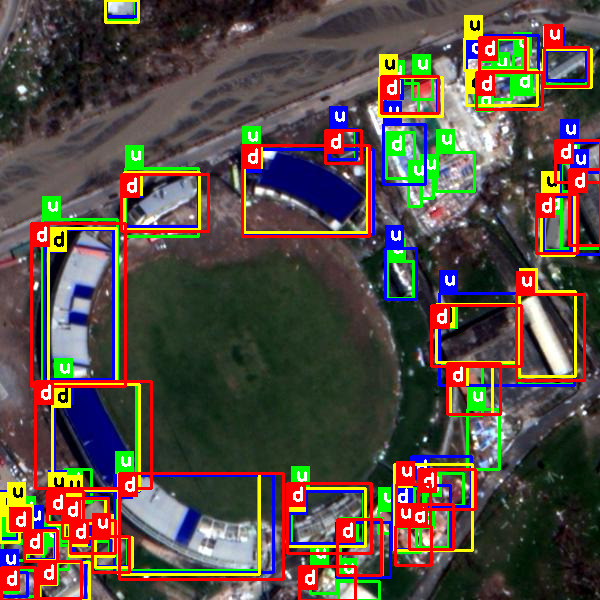}}
    \subcaptionbox{\label{fig:dis_example3}}{\includegraphics[width=0.19\linewidth,trim={6cm 0cm 0cm 10.7cm},clip]{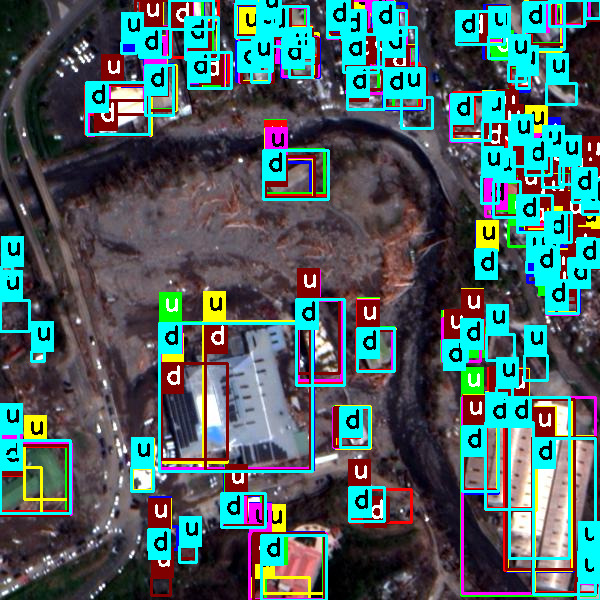}}
    \subcaptionbox{\label{fig:dis_example4}}{\includegraphics[width=0.19\linewidth,trim={0cm 6cm 8cm 6cm},clip]{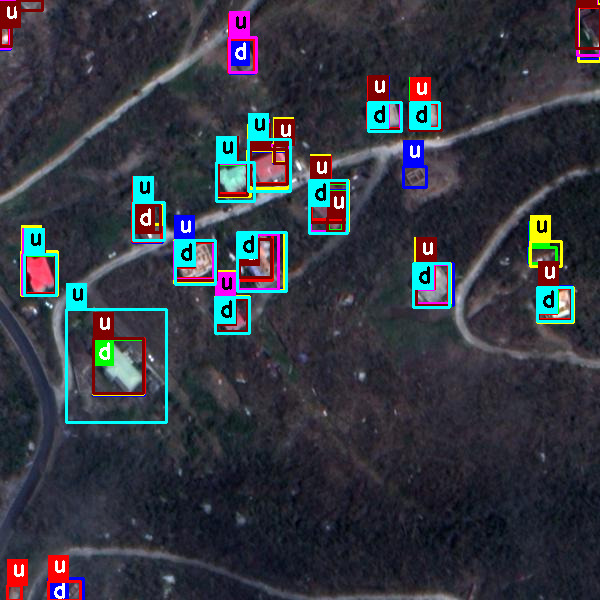}}
    \subcaptionbox{\label{fig:dis_example5}}{\includegraphics[width=0.19\linewidth,trim={7cm 11.5cm 0cm 0cm},clip]{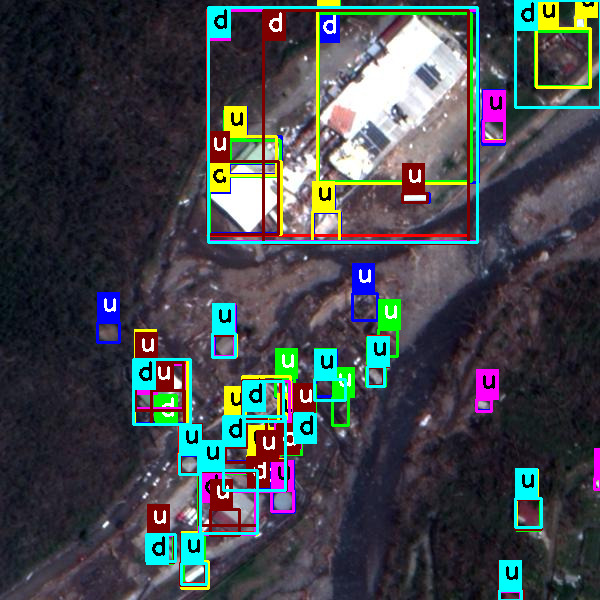}}
    \caption{Examples of ambiguous cases with noisy or incorrect annotations on (a - e) MS COCO \cite{cocodataset}, (f - j) VinDr-CXR \cite{vindr2022} and (k - o) disaster response dataset~\cite{bccnet2018} to identify damaged and undamaged buildings with class names `d' and `u', respectively. (a) Mislabelling one object with two bounding boxes. (b - d) Duplicate annotations of visually similar objects (\eg, car/truck, donut/cake and fork/spoon). (d)(e) Incorrect annotation of object classes (\eg, cake labelled as a bowl, only annotating one of the cake decorations as `person'). (f - o) Annotations from different annotators (represented by the different colours) of the VinDr-CXR and disaster response datasets show significant disagreement. (Details of the datasets are given in Section~\ref{sec:real_datasets}.)}
    \label{fig:example_hard}
\end{figure}

Despite significant advancements in the development of object detection with natural images \cite{fasterrcnn2015,wang2022yolov7,detr2020,eva2023}, existing methods typically assume the availability of large, accurately annotated datasets \cite{zhudataforobjdet, nowruzidataforobjdet}.
Nonetheless, this assumption does not often hold true in practice.
For example, accurate labels are often expensive to acquire in many applications.
A large number of experts with varying professional skills can be involved to crowdsource annotations of images sampled with replacement from a large dataset~\cite{agreetodisagree2019,crowdsourcingLena2014, canbudd2021}. Usually, the annotations from multiple annotators are unified and cleaned as the last step in annotation collection. However, disagreements arise when annotating complex images, making it challenging even for experienced expert annotators to achieve unanimity due to high interobserver variability \cite{mammointerobs2006, honeycombinterobs2012} and thus, rendering the role of an annotation arbitrator nearly impossible (see \Cref{fig:example_hard}).
As a result, \emph{multiple noisy annotations per image} are obtained. Such a setting for annotation crowdsourcing is often called the multi-rater problem. The majority of methods on multi-rater learning have been developed for image classification~\cite{agreetodisagree2019,yufusionbranch,jensenlabelsep} and segmentation \cite{probunet2018,jungolabelsep,phiseg2019,disgtmedimageseg2020,multiratersegWei2021,mrprism2021}, while we consider object detection here.

Several previous methods that exploit crowdsourced annotations have been proposed to improve object detection performance \cite{crowdrcnn2020,Le2023}. A common approach estimates the ground truth from noisy annotations through majority voting \cite{majorvote2008} which implicitly assumes that all annotators are equally accurate in their annotations. Alternatively, in \cite{Le2023}, the weight of each annotation is predicted based on the annotators' proficiency levels which are assumed to be known a priori; however, this information is difficult to obtain in practice. On the other hand, in \cite{crowdrcnn2020}, their proposed method is designed specifically for Faster R-CNN (FRCNN) \cite{fasterrcnn2015} and cannot be generalised to other object detectors. Moreover, these methods have been developed and evaluated on distinct, private synthetic datasets constructed using different methods. Thus, we cannot directly compare their reported experiment results.

\begin{figure}[tb]
     \centering
     \includegraphics[width=.95\textwidth]{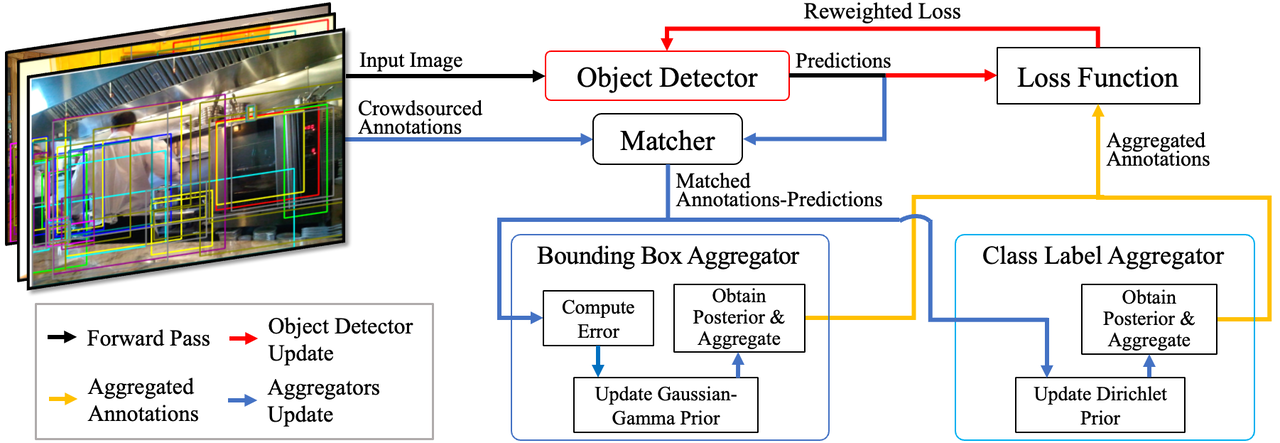}
    \caption{Overall architecture of the proposed BDC. 
    The updating of the aggregator's parameter and the object detector's parameters is repeated iteratively until convergence. Best viewed in colour.}
    \label{fig:overall_arch}
\end{figure}

To address these algorithmic and evaluation limitations, in this work, we first propose a model-agnostic annotation aggregation and object detector training pipeline, named \textit{Bayesian Detector Combination} (BDC).
BDC aggregates noisy bounding boxes and class labels by modelling the annotation reliability of each crowd member for both elements (\Cref{fig:overall_arch}), without requiring any additional knowledge of the ground truth annotations or their quality. Furthermore, due to the limited access to the ground truth of real crowdsourced datasets, we synthesise multiple datasets simulating various crowdsourcing scenarios for systematic evaluation of the previous works and the proposed method. The main contributions of our work include:

\begin{itemize}
    \item We propose a model-agnostic Bayesian Detector Combination (BDC) framework that simultaneously infers the annotation quality of each crowd member and infers consensus bounding boxes and soft class labels from noisy, crowdsourced annotations to improve model performance.
    \item We implement and systematically evaluate the previous methods on real-world and synthetic datasets with crowdsourced annotations which simulate varying crowdsourcing scenarios.
    \item Extensive experiments conducted demonstrate the superior performance of BDC compared with prior alternatives in key metrics such as the average precision, the scalability with the increasing number of annotators, and the robustness against noisy annotators.
\end{itemize}

%% file: sec/2_lit_review.tex
\section{Related Work}
\label{sect:litrev}

\paragraph{\bf Classification with Crowdsourced Annotations.} Learning with crowdsourced annotations for classification tasks has been studied in various works. Sheng \etal~\cite{majorvote2008} adopted a simple approach to aggregate multiple labels via majority voting (MV) where the mode class is selected as the ground truth. 
Several works \cite{davidskene1979, whitehillglad, Raykar2010} modelled the sensitivity and specificity of different annotators via confusion matrices obtained through the Expectation-Maximisation algorithm or a Bayesian approach.
Kim and Ghahramani~\cite{BCC2012} introduced a general framework for Bayesian classifier combination (BCC) to model the relationship between the unknown true label and the annotated class label which is extended by Venanzi \etal~\cite{commbcc2014} with the concept of annotator communities. 
Albarqouni \etal~\cite{aggnet2016} and Isupova \etal~\cite{bccnet2018} extended the idea of previous works \cite{Raykar2010,BCC2012} to convolutional neural networks (CNN) with softmax outputs. 
Several recent works use a trained classifier to rank consensus labels and incorporate the ranking scores into model learning \cite{gohcrowdlab, activelabel}. Another research direction is to take all noisy annotations separately as the input during the training process \cite{weilabelsep, chenlabelsep, guanlabelsep,jensenlabelsep,crowdlayer2018, yufusionbranch,agreetodisagree2019}, but Wei \etal~\cite{weilabelsepconc} concluded that label aggregation is preferred over label separation when there are sufficient number of annotators or the noise rate is moderate. These methods are not directly applicable to object detection tasks since they do not involve regressing the locations of bounding boxes.

\paragraph{\bf Object Detection with Crowdsourced Annotations.} Hu and Meina~\cite{crowdrcnn2020} proposed Crowd R-CNN where one first pre-processes the annotations by combining the bounding boxes using a clustering algorithm, before aggregating the labels and learning the sensitivity of each annotator \cite{Raykar2010} to different object classes during model training. 
Crowd R-CNN is tailored for Faster R-CNN (FRCNN) \cite{fasterrcnn2015} and is incompatible with one-stage object detectors such as YOLOv7 \cite{wang2022yolov7}.
Le \etal~\cite{Le2023} proposed to pre-process annotations using Weighted Boxes Fusion (WBF) \cite{WBF2021} and introduced expert agreement re-weighted loss (EARL) to emphasise boxes with higher annotator agreement for model training. It requires access to each annotator's proficiency level to perform WBF and compute EARL, which is often challenging to obtain for real-world crowdsourced data. Compared with the aforementioned methods, the proposed BDC framework can incorporate any object detector in a plug-and-play fashion without requiring any additional inputs such as annotators' proficiency levels. 
Furthermore, previous works \cite{crowdrcnn2020, Le2023} evaluated their methods on different synthetic datasets, with different numbers of annotators and proficiency levels. For example, Hu and Meina~\cite{crowdrcnn2020} performed evaluation on synthetic datasets derived from PASCAL VOC \cite{pascal-voc-2007} and MS COCO \cite{cocodataset}; Le \etal~\cite{Le2023} synthesised a MED-MNIST dataset based on MNIST \cite{mnist2010} and adopted the VinDr-CXR dataset~\cite{vindr2022}. Thus, their experiment results cannot be compared directly.

%% file: sec/3_methodology.tex
\section{Method}
\label{sect:method}

{\bf Problem Statement.} We assume a training set $D = \{x^i, y^i\}_{i=1}^N$ with pairs of input images $x^i$ and crowdsourced annotations $y^i = \{y^i_m\}_{m=1}^{M^i}$, where $N$ is the total number of training samples and $M^i$ is the number of annotations of the $i$-th image from all annotators.
Each annotation is a tuple of a bounding box, class label and the annotator ID, $y^i_m = \{(b^i_m, c^i_m, k^i_m)\}$ in which $b^i_m \in \mathbb{R}^4$ contains the x-y coordinates of the box centre as well as the width and height of the bounding box, $c^i_m \in \{1,...,J\}$ represents the annotated class label of the object and $k^i_m \in \{1,...,K\}$ is the annotator ID. $J$ denotes the number of object classes while $K$ is the total number of annotators. Note that each annotator may annotate a varying number of object instances. For brevity, notations in the rest of this section are simplified to a single training sample $i$. 
Our goal is to learn an object detector with noisy crowdsourced annotations while inferring and leveraging annotators' varying annotation abilities.

\subsection{Bayesian Detector Combination}
An overview of the proposed crowdsourced annotation aggregation framework is shown in \Cref{fig:overall_arch}. At a high level, the proposed framework consists of four components, namely: the object detector (OD) module, the annotations-predictions matcher, the bounding box aggregator (BBA) and the class label aggregator (CLA). Firstly, we perform a one-to-many matching of the predictions generated by the OD to the crowdsourced annotations. 
The bounding box error is computed and used to update a Gaussian-Gamma prior distribution of the BBA for each annotator. 
Next, the class probability predictions by the OD and the class label annotations are used to update a Dirichlet prior in the CLA. The posterior distributions obtained from both aggregators are used to aggregate the bounding boxes and class labels.
Finally, we train and optimise the OD's parameters and aggregators' parameters iteratively until convergence. We present a detailed introduction of each component below.

\subsubsection{Object Detector Module.}
\label{sec:od}
The object detector (OD) module with a parameter set, $\theta_{O}$, outputs a varying number of $P$ predictions, $\hat{y} = \{(\hat{b}_n, \hat{p}_n)\}_{n=1}^{P}$, given an input image, $x$, where $\hat{b}_n \in \mathbb{R}^4$  and $\hat{p}_n \in [0,1]^J$ are the predicted bounding box and class probability, respectively. %
We train the model 
by minimising an overall loss function:
\begin{equation}
    \mathcal{L_D} = \frac{1}{H} \sum^{H}_{n=1} \gamma_n \mathcal{L}(\mathfrak{b}_n, \rho_n, \hat{y})\,\,,
    \label{eq:od_loss}
\end{equation}
where $\gamma_n$, $\mathfrak{b}_n$ and $\rho_n$ are the loss scale factor, the aggregated bounding box, and class probabilities obtained from the two aggregators defined in \Cref{sec:BBA,sec:CLA}, respectively. $H$ is the number of aggregated annotations. $\mathcal{L}$ is the loss function that the base OD model adopts.
Since the OD operates as a black-box module in the proposed framework, any object detector can be used here as long as it outputs the predicted bounding boxes and the associated class label probabilities. 

\subsubsection{Annotations-Predictions Matcher.}
\label{sec:matcher}

To update the priors of the aggregators, we first match each annotation to one of the predictions. To find the optimal prediction, $\hat{y}_m^*$, for each annotation, $y_m$, we minimise a matching cost function:
\begin{equation}
    \hat{y}_m^* = \argmin_{\hat{y}_n \in \hat{y}} \mathcal{L}_{match}(\hat{y}_n, y_m)\,\,,
\end{equation}
where
\begin{equation}
\label{eq:matching_cost}
    \mathcal{L}_{match}(\hat{y}_n, y_m) = -\hat{p}_{n(c_m)} + \lambda_1 \mathcal{L}_{IoU}(\hat{b}_n, b_m) + \lambda_2 ||\hat{b}_n - b_m||_1\,\,.
\end{equation}
$\lambda_1, \lambda_2 \in \mathbb{R}$ are hyperparameters and $\mathcal{L}_{IoU}$ is the IoU loss \cite{giou}. In short, for each annotation, we aim to find the prediction with a higher predicted probability of the annotated class label, a higher IoU and a smaller L1 distance between the predicted and the annotated bounding box. This is inspired by the set prediction loss in DETR \cite{detr2020}. 

Instead of finding a one-to-one matching with a global minimum matching cost, we are interested in a one-to-many matching that one prediction can be matched to many crowdsourced annotations, with a local minimum matching cost, \ie finding the prediction with minimum cost for each annotation individually. Moreover, a simple heuristic matching rule is sufficient instead of the Hungarian algorithm in DETR which is more computationally demanding. 

\subsubsection{Bounding Box Aggregator.}
\label{sec:BBA}

In the Bounding Box Aggregator (BBA), the translation and scaling errors in the x and y-axis of each annotator are modelled with a Gaussian distribution:

\begin{equation}
    p(\epsilon_m | k_{m}=k, \bm{\mu}, \bm{\sigma}) = \mathcal{N}({\mu}^k, {\sigma}^k)\,\,,
    \label{eq:box_error_posterior}
\end{equation}
where $\epsilon_m \in \mathbb{R}^4$ contains the translation and scaling errors in the x and y-axis. $\bm{\mu}$ and $\bm{\sigma}$ are a collection of Gaussian parameters with ${\mu}^k$ and ${\sigma}^k$ representing the mean and covariance of $\epsilon_m$ for the annotator $k$, respectively. The Gaussian distributions have a Gaussian-Gamma prior with hyperparameters $\mu^k_0$, $\upsilon^k_0$ and $\beta^k_0$ \cite{gaussgamma,normalconj}. To update the prior distribution, we compute the four error terms between all matched prediction and annotation boxes for each annotator with:
\begin{equation}
    \epsilon_m = \Big[
    \begin{matrix}
        \hat{b}_{m(1)}^* - b_{m(1)}, & &
        \hat{b}_{m(2)}^* - b_{m(2)}, & &
        \hat{b}_{m(3)}^* \div b_{m(3)}, & &
        \hat{b}_{m(4)}^* \div b_{m(4)} 
    \end{matrix} 
    \Big]\,\,,
    \label{eq:box_error}
\end{equation}
where $ \hat{b}_m^*$ and $b_m$ are the matched prediction and annotation boxes, respectively. The subtraction and division operations are designed to make the error terms translation-invariant and scale-invariant. The Gaussian-Gamma prior is updated as follows:

\noindent\begin{minipage}{.49\linewidth}
    \begin{equation}
    \Tilde{\mu}^k = \frac{\mu^k_0 + M^k \bar{\epsilon}^k}{M^k + 1}\,\,,
    \label{eq:mu_update_main}
\end{equation}
\end{minipage}
\begin{minipage}{.5\linewidth}
    \begin{equation}
    \Tilde{\upsilon}^k = \upsilon^k_0 + \frac{M^k}{2}\,\,,
    \label{eq:ups_update_main}
\end{equation}
\end{minipage}

\begin{equation}
    \Tilde{\beta}^k = \beta^k_0 + \frac{M^k(\bar{\epsilon}^k - \mu^k_0)(\bar{\epsilon}^k - \mu^k_0)^\top}{2(M^k + 1)} + \frac{1}{2}\sum_{m=1}^{M} \mathbbm{1}[k_m=k](\epsilon_m - \bar{\epsilon}^k)(\epsilon_m - \bar{\epsilon}^k)^\top\,\,,
    \label{eq:beta_update_main}
\end{equation}
where $M^k$ is the number of annotation-prediction matches associated with the $k$-th annotator, $\bar{\epsilon}^k$ is the mean of all $\epsilon_m$ for the $k$-th annotator and $\mathbbm{1}[\cdot]$ is the indicator function.

The obtained posterior mean (\ie $\mu^k = \Tilde{\mu}^k$) is used to correct the annotated boxes for each annotator:
\begin{equation}
    b_m := (b_m 
    + 
    \begin{bmatrix}
        \mu^k_{(1)}, & \mu^k_{(2)}, & 0, & 0
    \end{bmatrix})
    \odot
    \begin{bmatrix}
        1, & 1, & \mu^k_{(3)}, & \mu^k_{(4)}
    \end{bmatrix}\,.
    \label{eq:correct_box}
\end{equation}
All annotations matched to the same prediction are aggregated via a weighted average using the posterior precision, $P^k=1 \oslash diag(\sigma^k)$, where $\oslash$ denotes the element-wise division and $\sigma^k = \Tilde{\beta}^k / \Tilde{\upsilon}^k$. The posterior precision, $P^k$, of each annotator is used as weights to obtain the final boxes, $\{\mathfrak{b}_n\}_{n=1}^H$, with
\begin{equation}
    \mathfrak{b}_n = \frac{\sum_{(b,k) \in \kappa_n} P^k \odot b}{\sum_{(b,k) \in \kappa_n} P^k}\,\,.
    \label{eq:aggregate_box}
\end{equation}
Here, 
$H$ indicates the number of predictions matched to at least one annotation which is also equal to the number of aggregated annotations defined in \Cref{sec:od}. 
$\kappa_n$ is the set of tuples consisting of the annotated bounding boxes and their annotator ID that are matched to the $n$-th prediction. Inspired by EARL \cite{Le2023}, a loss scale factor, $\gamma_n$, is computed for each aggregated box as the ratio of the number of matched annotations to $K$ and is used to re-weight the loss function of the OD in \Cref{eq:od_loss}. We refer the readers to Appendix~\ref{appendix:bba} for further details on the updating procedure of the Bounding Box Aggregator.

\subsubsection{Class Label Aggregator.}
\label{sec:CLA}
To aggregate crowdsourced class labels, we extend the Bayesian classifier combination neural network (BCCNet) \cite{bccnet2018} to the proposed BDC framework. In the class label aggregator (CLA), the annotated class label from each annotator, $c_m$, is modelled as a multinomial distribution conditioning on the true label of the object: 
\begin{equation}
    p(c_m | k_m = k, t_m = j, \bm{\pi}) = {\pi}^k_{j, c_m}\,\,,
\end{equation}
where $\bm{\pi}$ is a collection of confusion matrices, ${\pi}^k \in \mathbb{R}^{J \times J}$ is the confusion matrix of the $k$-th annotator and $t_m$ is the ground truth class label. Each row, ${\pi}^k_j$, provides the event probabilities of a multinomial distribution for the $k$-th annotator when $t_m=j$ and is modelled with a Dirichlet prior with hyperparameters ${\alpha}^k_{0j} = \{\alpha^k_{0j,1}, ..., \alpha^k_{0j,J}\}$ such that $\pi^k_j | \alpha^k_{0j} \sim Dir(\alpha^k_{0j})$ \cite{BCC2012}. 

The multinomial posterior distribution of the true label (\ie the aggregated class label probability) can be obtained as
\begin{equation}
    \rho_{n, j} = \exp \left(\ln \hat{p}_{n, j} + \sum_{(c,k) \in \Tilde{\kappa}_n} \mathbb{E}_{\pi^{k}_j}\ln \pi^{k}_{j, c}\right)\,\,,
    \label{eq:rho}
\end{equation}
where $\Tilde{\kappa}_n$ is the set of tuples consisting of the annotated class labels and their annotator ID that are matched to the $n$-th prediction. $\hat{p}_{n,j}$ is the OD's class-$j$ probability of the $n$-th prediction.
Note that the CLA outputs class probabilities (soft labels) unlike other methods (\eg MV) that output a single class (hard label).
\Cref{appendix:CLA} provides the derivation for Equation~\ref{eq:rho} and a detailed discussion of the updating steps of the Class Label Aggregator.

\subsubsection{Training.}
The optimal OD's parameters, $\theta_O$, and aggregators' parameters, $\bm{\mu}$, $\bm{\upsilon}$, $\bm{\beta}$ and $\bm{\alpha}$, are learnt iteratively in the BDC framework. Firstly, we initialise the parameters either randomly or with prior knowledge of the domain. Next, we train the OD by optimising the loss function, $\mathcal{L_D}$, with the aggregated annotations obtained using the current values of the aggregators' parameters. After one epoch of training, we obtain the OD's predictions using the current value of $\theta_O$ to update the prior distributions of aggregators. These two steps are repeated iteratively until the parameters converge. See Appendix~\ref{appendix:VB} for a further discussion of the training steps.

%% file: sec/4_5_exp_and_discuss.tex
\section{Experiments}

{\bf Training details.}
To demonstrate the generalisability and robustness of BDC in incorporating different object detection algorithms, three common object detection algorithms are used in our experiments, YOLOv7 \cite{wang2022yolov7}, FRCNN \cite{fasterrcnn2015} with the ResNet50 backbone and EVA \cite{eva2023}. During training, we randomly apply data augmentation such as rotation, translation, zooming and horizontal flipping to the training images and annotations. $\lambda_1, \lambda_2$ in \Cref{eq:matching_cost} are set to 2 and 5, respectively, following Carion \etal~\cite{detr2020}.
Lastly, the Gaussian-Gamma prior of the BBA, $\bm{\mu}_0$, $\bm{\upsilon}_0$ and $\bm{\beta}_0$, are set to 0, 10 and 0.5, respectively, while the Dirichlet prior of the CLA, $\bm{\alpha_0}$, is initialised to 1 and each diagonal entry is set to 10. The complete training configuration can be found in Appendix~\ref{appendix:hyperparams}.

\paragraph{\bf Comparison against other aggregation methods.}
We compare the performance of BDC against multiple baselines as shown in \Cref{table:vincxr_metrics,table:yolometrics,table:frcnnmetrics,table:detrmetrics}. `NA' denotes the process of training the model with all the crowdsourced annotations as the ground truth annotations \cite{weilabelsepconc}. `MV' stands for majority voting where we apply clustering to the crowdsourced annotations, then aggregate the class labels via MV and aggregate the bounding boxes as the biggest box where the majority had annotated pixel-wise, following the practice in \cite{majorvote2008}. We also implement Crowd R-CNN \cite{crowdrcnn2020} and WBF-EARL \cite{Le2023} designed for crowdsourced object detection. For WBF-EARL \cite{Le2023} which requires the knowledge of annotators' proficiency levels, we followed the settings in \cite{Le2023} for the VinDR-CXR dataset while for the synthetic datasets, they are set to the overall accuracy of the annotators.

\subsection{Real Crowdsourced Datasets}
\label{sec:real_datasets}
\paragraph{\bf VinDr-CXR.} We evaluate the annotation aggregation methods on VinDr-CXR \cite{vindr2022}, a thoracic abnormalities detection dataset. VinDr-CXR consists of 14 abnormalities classes and contains 15,000 training and 3,000 testing chest radiograph images. The training images are annotated by 17 experienced radiologists where at least three annotators are randomly assigned to annotate each image. \Cref{fig:vindr_example1,fig:vindr_example2,fig:vindr_example3,fig:vindr_example4,fig:vindr_example5} show examples of the annotations with high bounding box and class label disagreement by the annotators. Furthermore, the testing images are annotated with the consensus of five expert radiologists through a two-stage labelling process and the test set annotations are not publicly available. The predictions are submitted to the VinBigData Kaggle \cite{vindrkaggle} to obtain mean average precision (AP) at 0.4 IoU threshold 
on the private test set.

\paragraph{\bf Disaster Response Dataset.} We further use the disaster response dataset~\cite{bccnet2018} where the task is to detect undamaged and damaged buildings on satellite images taken after hurricane Maria in 2017. The dataset consisting of 464 satellite images of Dominica is annotated by 13 semi-professional paid crowd members to ensure high labelling quality. Despite this effort, labels from different crowd members vary significantly (\Cref{fig:dis_example1,fig:dis_example2,fig:dis_example3,fig:dis_example4,fig:dis_example5}). Consequently, the dataset does not contain ground truth labels. Thus, its qualitative results are presented in Appendix~\ref{appendix:disaster_response}.

\begin{table}[tb]
    \caption{AP$^{.4}$ results on VinDr-CXR \cite{vindr2022} dataset obtained from the private score of VinBigData Kaggle competition \cite{vindrkaggle}. Crowd R-CNN can only integrate FRCNN, hence there are no results reported for the two other entries corresponding to Crowd R-CNN.}
    \label{table:vincxr_metrics}
    \centering
    \setlength{\tabcolsep}{5pt}
    \resizebox{.53\columnwidth}{!}{
    \begin{tabular}{l|c|c|c}
        \hline
        \multirow{2}{*}{Method} & \multicolumn{3}{c}{Test AP$^{.4}$} \\
        & YOLOv7 & FRCNN & EVA \\
        \hline
        NA & 17.4 & 17.2 & 7.8 \\
        MV & 13.9 & 16.3 & 8.2 \\
        Crowd R-CNN \cite{crowdrcnn2020} & - & 16.7 & - \\
        WBF-EARL \cite{Le2023} & 16.4 & 17.0 & 8.4\\
        \textbf{BDC (ours)} & \textbf{19.2} & \textbf{17.9} & \textbf{8.9}\\
        \hline
    \end{tabular}
    }
\end{table}

\paragraph{\bf Quantitative Results.} \Cref{table:vincxr_metrics} presents the AP$^{.4}$ metric obtained from the VinBigData Kaggle competition and shows that the proposed BDC leads to higher AP$^{.4}$ than other methods across different OD models on the VinDr-CXR test set. 
As Crowd R-CNN can only integrate FRCNN, there are no results reported for the two other entries corresponding to Crowd R-CNN.

\begin{figure}[tb]
    \captionsetup[subfigure]{format=hang}
    \begin{subfigure}{0.19\textwidth}
        \centering \includegraphics[trim={0 1cm 0 4.2cm}, clip,width=\linewidth]{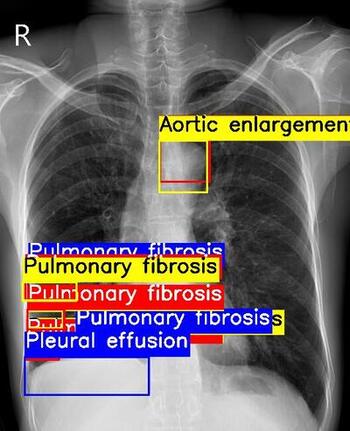 }
    \end{subfigure}
    \begin{subfigure}{0.19\textwidth}
        \centering \includegraphics[trim={0 1cm 0 4.2cm}, clip,width=\linewidth]{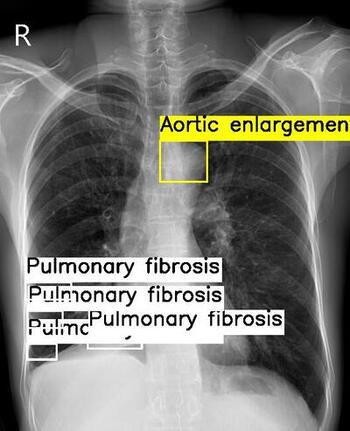}
    \end{subfigure}
    \begin{subfigure}{0.19\textwidth}
        \centering \includegraphics[trim={0 1cm 0 4.2cm}, clip,width=\linewidth]{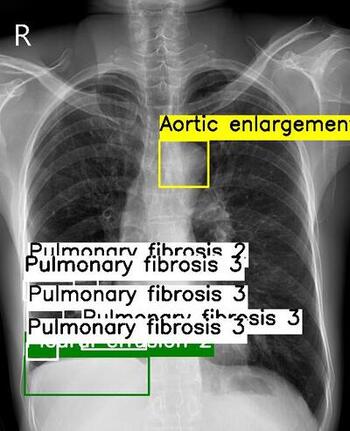}
    \end{subfigure}
    \begin{subfigure}{0.19\textwidth}
        \centering \includegraphics[trim={0 1cm 0 4.2cm}, clip,width=\linewidth]{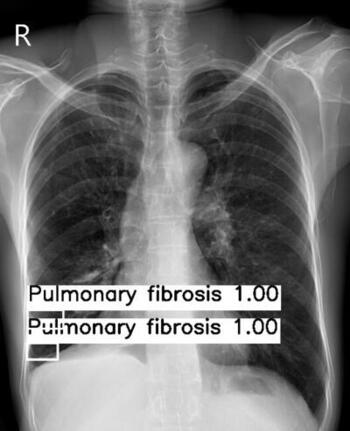}
    \end{subfigure}
    \begin{subfigure}{0.19\textwidth}
        \centering \includegraphics[trim={0 1cm 0 4.2cm}, clip,width=\linewidth]{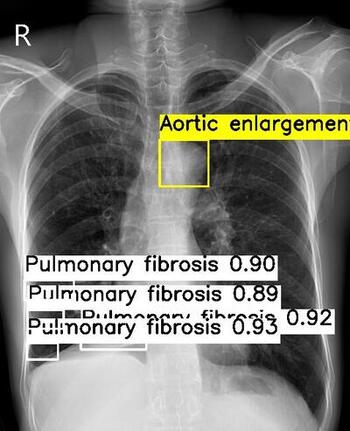}
    \end{subfigure}
    \medskip
    
    \subcaptionbox{NA}{\includegraphics[trim={0 3.8cm 0 2cm}, clip,width=0.19\linewidth]{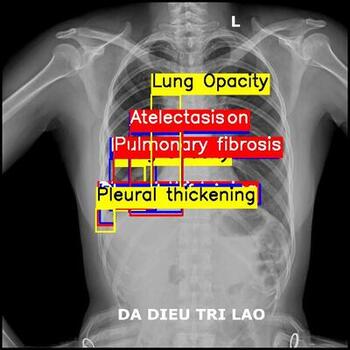}}
    \subcaptionbox{MV}{\includegraphics[trim={0 3.8cm 0 2cm}, clip,width=0.19\linewidth]{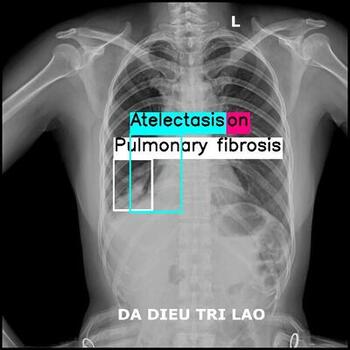}}
    \subcaptionbox{WBF-EARL}{\includegraphics[trim={0 3.8cm 0 2cm}, clip,width=0.19\linewidth]{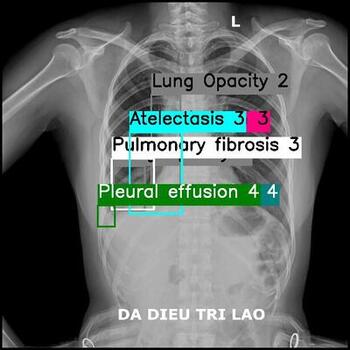}}
    \subcaptionbox{Crowd R-CNN}{\includegraphics[trim={0 3.8cm 0 2cm}, clip,width=0.19\linewidth]{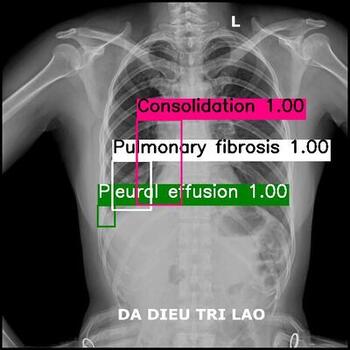}}
    \subcaptionbox{BDC (ours)}{\includegraphics[trim={0 3.8cm 0 2cm}, clip,width=0.19\linewidth]{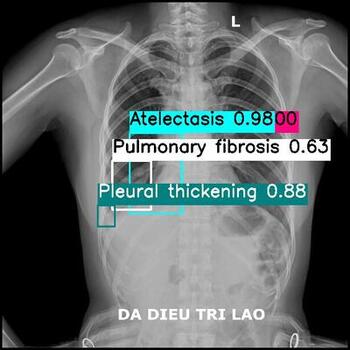}}

    \caption{Comparison of aggregated labels from different methods on the VinDr-CXR \cite{vindr2022}. For WBF-EARL \cite{Le2023}, the number beside the class label is the annotators' level of agreement while for Crowd R-CNN \cite{crowdrcnn2020} and BDC, the number indicates the class probability. For NA, the colours represent the different annotators.}
    \label{fig:qualitative_real}
\end{figure}

\paragraph{\bf Qualitative Results.} We visualise the results of the aggregated annotations from each method on the VinDr-CXR dataset in \Cref{fig:qualitative_real}. Since the true ground truth of the datasets is not provided, we use `MV' as the best approximation of the ground truth for this qualitative assessment and observe that BDC produces more accurate aggregation compared to the other methods.

\subsection{Synthetic Datasets}
\label{sec:syn_datasets}
To alleviate the limitations in the available real-world crowdsourced datasets including the lack of ground truth labels, we synthesise several synthetic datasets with crowdsourced annotations to enable the systematic evaluation of the proposed method. This makes available the ground truth annotations to allow us to obtain detailed test evaluation metrics (e.g., AP$^{.5}$, AP$^{.75}$, AP$^{.5:.95}$) and verify the performance of the proposed method by evaluating the aggregated crowdsourced annotations with the true ground truth annotations of the training dataset (indicated as `Train Aggregation AP' in \Cref{table:yolometrics,table:frcnnmetrics,table:detrmetrics}). Two common object detection datasets are used to construct the synthetic datasets: (1) Pascal Visual Object Classes (VOC) 2007 \cite{pascal-voc-2007}, a 20-class object detection dataset which consists of 5,011 and 4,952 images for training and testing, respectively; (2) Microsoft Common Objects in Context (COCO) 2017 \cite{cocodataset}, a large-scale 80-class object detection dataset consisting of 118,287 training and 5,000 validation images.

\paragraph{\bf Synthetic settings.}
Four synthetic settings are investigated in this work: (1) \emph{VOC-FULL} and (2) \emph{COCO-FULL} where 10 annotators with average skill levels annotate the whole datasets; (3) \emph{VOC-MIX} where the first 5 annotators are experts while the remaining 20 annotators have an average skill level. All annotators annotate the whole dataset since Pascal VOC is a small dataset; and (4) \emph{COCO-MIX} where 20 experts, 550 annotators with average skill level and 330 annotators with poor skill level work together to annotate the MS COCO dataset. Additionally, the last 100 are random annotators, \ie, the annotated class labels and bounding boxes are drawn from a uniform distribution. Since MS COCO is a large-scale dataset, we set that each annotator annotates a small subset of images, \ie, the coverage for the expert, average, poor and random annotators are 5\%, 1\%, 0.5\% and 1\%, respectively. 
Note that all annotators in these settings are simulated using deep learning models (See Appendix~\ref{appendix:pseudocode_synthesise} for further discussions of the synthesisation method and pseudocode).
Additional experiments where we directly used the object detectors' predictions as synthetic crowdsourced annotations can be found in \Cref{appendix:bdc_extension}.

\begin{table}[tb]
    \caption{Test and train aggregation AP results for YOLOv7 model trained with different annotation aggregation methods on synthesised and ground truth datasets.}
    \label{table:yolometrics}
    \centering
    \setlength{\tabcolsep}{5pt}
    \resizebox{.83\columnwidth}{!}{
        \begin{tabular}{l|l||ccc|ccc}
            \hline
             \multirow{2}{*}{Dataset} & \multirow{2}{*}{Method} & \multicolumn{3}{c|}{Test AP} & \multicolumn{3}{c}{Train Aggregation AP} \\
             & & AP$^{.5}$ & AP$^{.75}$ & AP$^{.5:.95}$ & AP$^{.5}$ & AP$^{.75}$ & AP$^{.5:.95}$ \\
             \hline
              \multirow{4}{*}{\makecell[l]{VOC-\\FULL}} & NA & 59.7 & 45.3 & 40.7 & 10.2 & 9.9 & 9.9 \\
              & MV & 72.2 & 52.3 & 47.3 & 62.0 & 48.5 & 48.5 \\
              & WBF-EARL \cite{Le2023} & 69.3 & 52.3 & 45.7 & 22.3 & 16.4 & 14.5 \\
              & \textbf{BDC (ours)} & \textbf{84.9} & \textbf{63.5} & \textbf{54.8} & \textbf{95.0} & \textbf{80.6} & \textbf{66.3} \\
              \hline
              \multirow{4}{*}{\makecell[l]{VOC-\\MIX}} & NA & 46.1 & 27.6 & 27.4 & 4.4 & 4.3 & 4.3 \\
              & MV & 77.1 & 61.0 & 53.7 & 75.7 & 70.6 & 68.6 \\
              & WBF-EARL \cite{Le2023}& 61.0 & 47.2 & 40.9 & 27.9 & 24.9 & 20.0 \\
              & \textbf{BDC (ours)} & \textbf{86.2} & \textbf{69.0} & \textbf{58.8} & \textbf{97.6} & \textbf{94.9} & \textbf{77.2} \\
              \hline
              VOC & GT & 87.7 & 74.0 & 65.3 & 100 & 100 & 100 \\
              \hline
              \hline
              \multirow{4}{*}{\makecell[l]{COCO-\\FULL}} & NA & 53.4 & 41.4 & 37.4 & 8.6 & 8.2 & 8.2\\
              & MV & 61.9 & 46.4 & 42.7 & 52.5 & 44.4 & 43.3 \\
              & WBF-EARL \cite{Le2023}& 55.6 & 42.5 & 38.9 & 14.6 & 8.9 & 8.6 \\
              & \textbf{BDC (ours)} & \textbf{65.0} & \textbf{49.6} & \textbf{44.5} & \textbf{83.6} & \textbf{66.1} & \textbf{56.9} \\
              \hline
              \multirow{4}{*}{\makecell[l]{COCO-\\MIX}} & NA & 53.4 & 41.2 & 37.5 & 8.5 & 7.9 & 8.0 \\
              & MV & 60.9 & 46.4 & 42.4 & 45.8 & 42.3 & 41.1 \\
              & WBF-EARL \cite{Le2023}& 56.2 & 42.6 & 39.1 & 12.1 & 7.0 & 7.0 \\
              & \textbf{BDC (ours)} & \textbf{64.8} & \textbf{49.3} & \textbf{44.4} & \textbf{80.5} & \textbf{62.9} & \textbf{54.3} \\
              \hline
              COCO & GT & 67.5 & 52.8 & 48.6 & 100 & 100 & 100 \\
              \hline
        \end{tabular}
    }
\end{table}

\paragraph{\bf Quantitative Results.}

\begin{figure}[tb]
    \captionsetup[subfigure]{format=hang}
    \begin{subfigure}{0.16\textwidth}
        \centering \includegraphics[trim={0 30pt 0 0},width=\linewidth]{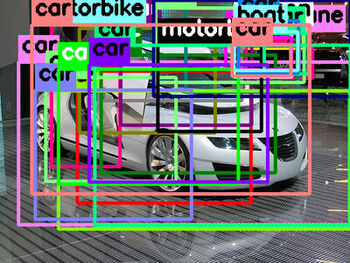}
    \end{subfigure}
    \begin{subfigure}{0.16\textwidth}
        \centering \includegraphics[trim={0 30pt 0 0},width=\linewidth]{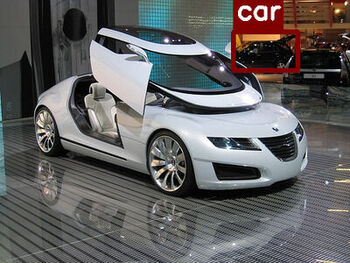}
    \end{subfigure}
    \begin{subfigure}{0.16\textwidth}
        \centering \includegraphics[trim={0 30pt 0 0},width=\linewidth]{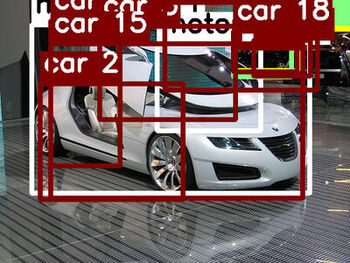}
    \end{subfigure}
    \begin{subfigure}{0.16\textwidth}
        \centering \includegraphics[trim={0 30pt 0 0},width=\linewidth]{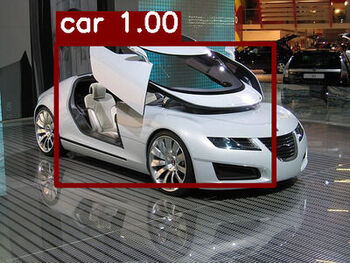}
    \end{subfigure}
    \begin{subfigure}{0.16\textwidth}
        \centering \includegraphics[trim={0 30pt 0 0},width=\linewidth]{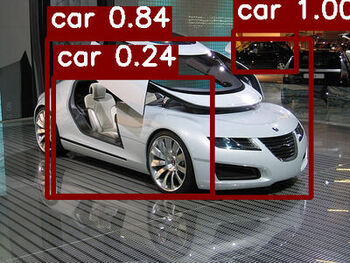}
    \end{subfigure}
    \begin{subfigure}{0.16\textwidth}
        \centering \includegraphics[trim={0 30pt 0 0},width=\linewidth]{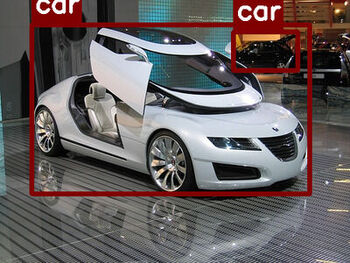}
    \end{subfigure}

    \medskip
    \subcaptionbox{NA}{\includegraphics[width=0.16\linewidth]{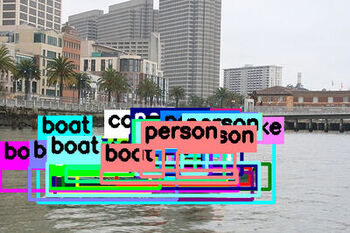}}
    \subcaptionbox{MV}{\includegraphics[width=0.16\linewidth]{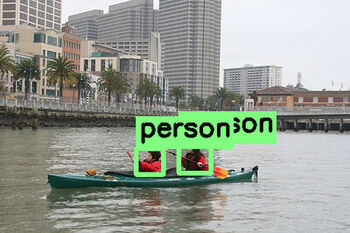}}
    \subcaptionbox{WBF-EARL}{\includegraphics[width=0.16\linewidth]{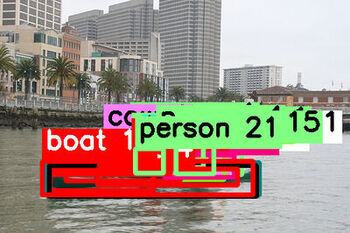}}
    \subcaptionbox{Crowd R-CNN}{\includegraphics[width=0.16\linewidth]{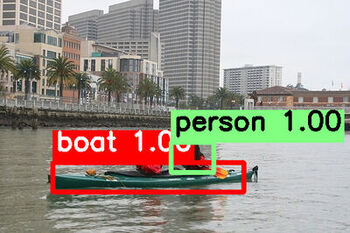}}
    \subcaptionbox{BDC (ours)}{\includegraphics[width=0.16\linewidth]{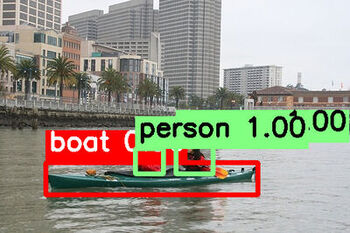}}
    \subcaptionbox{Ground truth}{\includegraphics[width=0.16\linewidth]{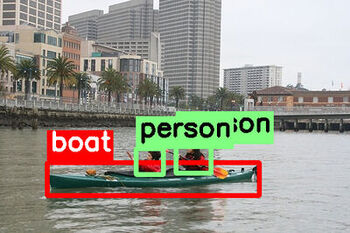}}
    \caption{Comparison of aggregated labels from different methods on the VOC-MIX synthetic dataset. For WBF-EARL \cite{Le2023}, the number beside the class label is the annotators' level of agreement while for Crowd R-CNN \cite{crowdrcnn2020} and BDC, the number indicates the class probability. For NA, the colours represent the different annotators.}
    \label{fig:qualitative}
\end{figure}

\Cref{table:yolometrics,table:frcnnmetrics,table:detrmetrics} report the experiment results of YOLOv7, FRCNN and EVA trained with different aggregation methods on all synthetic settings, respectively. Additionally, we trained each model with ground truth annotations of each dataset (denoted by `GT') to obtain reference optimal model performance. 
The aggregation APs of the train dataset for NA and WBF-EARL are low because NA directly uses all the crowdsourced annotations while WBF outputs multiple bounding boxes per class per object instance, leading to a much lower precision (see \Cref{fig:qualitative} for examples). We do not include performance from Crowd R-CNN for the COCO-MIX setting as its training time increases linearly with $K$ which means that the training time is prohibitively high for this dataset where $K=1,000$. In comparison, the extra computational cost introduced by the proposed BDC is small when compared to the much heavier training cost of the OD backbone (see \Cref{sec:time_analysis} for an analysis of the computational cost). \Cref{table:yolometrics,table:frcnnmetrics,table:detrmetrics} demonstrate the superior performance of BDC compared to other methods from simpler synthetic settings to more complex synthetic crowdsourced annotations, \ie from VOC-FULL to COCO-MIX. In addition, BDC achieves the highest train aggregation AP compared to other methods across all settings.

\begin{table}[tb]
    \caption{Test and train aggregation AP results for FRCNN model trained with different annotation aggregation methods on synthesised and ground truth datasets. Crowd R-CNN has a training time that increases linearly with $K$, hence there is no result reported for COCO-MIX dataset with $K=1,000$.}
    \label{table:frcnnmetrics}
    \centering
    \setlength{\tabcolsep}{5pt}
    \resizebox{0.87\columnwidth}{!}{
        \begin{tabular}{l|l||ccc|ccc}
            \hline
             \multirow{2}{*}{Dataset} & \multirow{2}{*}{Method} & \multicolumn{3}{c|}{Test AP} & \multicolumn{3}{c}{Train Aggregation AP} \\
             & & AP$^{.5}$ & AP$^{.75}$ & AP$^{.5:.95}$ & AP$^{.5}$ & AP$^{.75}$ & AP$^{.5:.95}$ \\
             \hline
              \multirow{5}{*}{\makecell[l]{VOC-\\FULL}} & NA & 47.4 & 11.3 & 19.4 & 10.2 & 9.9 & 9.9 \\
              & MV & 79.7 & 53.5 & 48.8 & 62.0 & 48.5 & 48.5 \\
              & Crowd R-CNN \cite{crowdrcnn2020} & 71.6 & 33.9 & 30.1 & 58.3 & 43.0 & 31.5 \\
              & WBF-EARL \cite{Le2023}& 71.8 & 37.7 & 38.7 & 22.3 & 16.4 & 14.5 \\
              & \textbf{BDC (ours)} & \textbf{83.7} & \textbf{56.3} & \textbf{51.0} & \textbf{95.0} & \textbf{80.4} & \textbf{66.3} \\
              \hline
              \multirow{5}{*}{\makecell[l]{VOC-\\MIX}} & NA & 34.1 & 4.4 & 12.2 & 4.4 & 4.3 & 4.3 \\
              & MV & 82.0 & 57.1 & 51.9 & 75.7 & 70.6 & 68.6 \\
              & Crowd R-CNN \cite{crowdrcnn2020}& 70.2 & 31.2 & 29.3 & 60.6 & 44.2 & 31.0\\
              & WBF-EARL \cite{Le2023}& 69.8 & 36.5 & 37.4 & 27.9 & 24.9 & 20.0 \\
              & \textbf{BDC (ours)} & \textbf{84.7} & \textbf{59.3} & \textbf{52.9} & \textbf{97.6} & \textbf{94.6} & \textbf{76.9} \\
              \hline
              VOC & GT & 84.6 & 62.1 & 55.2 & 100 & 100 & 100 \\
              \hline
              \hline
              \multirow{5}{*}{\makecell[l]{COCO-\\FULL}} & NA & 39.7 & 12.0 & 17.7 & 8.6 & 8.2 & 8.2 \\
              & MV & 55.6 & \textbf{35.7} & 33.4 & 52.5 & 44.4 & 43.3 \\
              & Crowd R-CNN \cite{crowdrcnn2020} & 48.5 & 27.1 & 24.2 & 49.5 & 36.2 & 33.7 \\
              & WBF-EARL \cite{Le2023}& 51.9 & 28.0 & 28.4 & 14.6 & 8.9 & 8.6 \\
              & \textbf{BDC (ours)} & \textbf{56.6} & 35.3 & \textbf{33.5} & \textbf{83.3} & \textbf{66.0} & \textbf{56.7} \\
              \hline
              \multirow{5}{*}{\makecell[l]{COCO-\\MIX}} & NA & 43.9 & 17.5 & 21.4 & 8.5 & 7.9 & 8.0 \\
              & MV & 54.8 & 34.5 & 32.9 & 45.8 & 42.3 & 41.1 \\
              & Crowd R-CNN \cite{crowdrcnn2020} & - & - & - & - & - & - \\
              & WBF-EARL \cite{Le2023}& 51.9 & 28.0 & 28.3 & 12.1 & 7.0 & 7.0 \\
              & \textbf{BDC (ours)} & \textbf{56.1} & \textbf{34.6} & \textbf{33.0} & \textbf{80.3} & \textbf{62.7} & \textbf{54.1} \\
              \hline
              COCO & GT & 58.0 & 38.5 & 35.7 & 100 & 100 & 100 \\
              \hline
        \end{tabular}
    }
\end{table}

\begin{table}[tb]
    \caption{Test and train aggregation AP results for EVA model trained with different annotation aggregation methods on synthesised and ground truth datasets.}
    \label{table:detrmetrics}
    \centering
    \setlength{\tabcolsep}{5pt}
    \resizebox{.83\columnwidth}{!}{
        \begin{tabular}{l|l||ccc|ccc}
            \hline
             \multirow{2}{*}{Dataset} & \multirow{2}{*}{Method} & \multicolumn{3}{c|}{Test AP} & \multicolumn{3}{c}{Train Aggregation AP} \\
             & & AP$^{.5}$ & AP$^{.75}$ & AP$^{.5:.95}$ & AP$^{.5}$ & AP$^{.75}$ & AP$^{.5:.95}$ \\
             \hline
              \multirow{4}{*}{\makecell[l]{VOC-\\FULL}} & NA & 90.9 & 80.9 & 69.7 & 10.2 & 9.9 & 9.9 \\
              & MV & 91.0 & 82.2 & 71.2 & 62.0 & 48.5 & 48.5 \\
              & WBF-EARL \cite{Le2023} & 93.0 & 83.5 & 71.1 & 22.3 & 16.4 & 14.5 \\
              & \textbf{BDC (ours)} & \textbf{94.2} & \textbf{85.6} & \textbf{71.7} & \textbf{95.3} & \textbf{81.2} & \textbf{66.7} \\
              \hline
              \multirow{4}{*}{\makecell[l]{VOC-\\MIX}} & NA & 86.6 & 65.7 & 57.5 & 4.4 & 4.3 & 4.3 \\
              & MV & 91.7 & 83.0 & 72.9 & 75.7 & 70.6 & 68.6 \\
              & WBF-EARL \cite{Le2023} & 91.3 & 82.2 & 68.7 & 27.9 & 24.9 & 20.0 \\
              & \textbf{BDC (ours)} & \textbf{94.9} & \textbf{86.5} & \textbf{73.8} & \textbf{98.8} & \textbf{95.0} & \textbf{77.3} \\
              \hline
              VOC & GT & 94.9 & 88.1 & 78.0 & 100 & 100 & 100 \\
              \hline
              \hline
              \multirow{4}{*}{\makecell[l]{COCO-\\FULL}} & NA & 71.8 & 55.7 & 50.6 & 8.6 & 8.2 & 8.2 \\
              & MV & 74.8 & 61.0 & 54.4 & 52.5 & 44.4 & 43.3 \\
              & WBF-EARL \cite{Le2023} & 74.7 & 58.4 & 51.8 & 14.6 & 8.9 & 8.6 \\
              & \textbf{BDC (ours)} & \textbf{78.0} & \textbf{62.7} & \textbf{55.1} & \textbf{83.3} & \textbf{65.1} & \textbf{55.3} \\
              \hline
              \multirow{4}{*}{\makecell[l]{COCO-\\MIX}} & NA & 73.3 & 57.2 & 51.7 & 8.5 & 7.9 & 8.0 \\
              & MV & 75.0 & 60.6 & 54.2 & 45.8 & 42.3 & 41.1 \\
              & WBF-EARL \cite{Le2023} & 75.2 & 59.3 & 52.9 & 12.1 & 7.0 & 7.0 \\
              & \textbf{BDC (ours)} & \textbf{77.1} & \textbf{62.5} & \textbf{54.8} & \textbf{79.9} & \textbf{62.2} & \textbf{53.2} \\
              \hline
              COCO & GT & 78.8 & 64.9 & 58.8 & 100 & 100 & 100 \\
              \hline
        \end{tabular}
    }
\end{table}

\paragraph{\bf Qualitative Results.}
\Cref{fig:qualitative} visualises the results of the aggregated annotations from each method on the VOC-MIX synthetic dataset. WBF performs box fusion on a per-class basis and results in significantly more boxes compared to other methods (excluding NA) which contributed to the low train aggregation AP and noisy labels during model training. On the other hand, although both Crowd R-CNN and BDC output class probabilities, the probability distribution of Crowd R-CNN is heavily skewed which is similar to hard labels, while BDC outputs the learnt class probabilities and can benefit from these soft labels during model training \cite{softlabels}. 
Besides, BDC can cope with failure cases of its aggregator as shown by the lower class probability of incorrect aggregation. 
Additional qualitative results can be found in Appendix~\ref{appendix:add_qualitative}.

\subsection{Scalability and Robustness of BDC} 
\begin{figure}[tb]
    \centering
    \begin{subfigure}{\textwidth}
        \centering\includegraphics[width=0.6\linewidth]{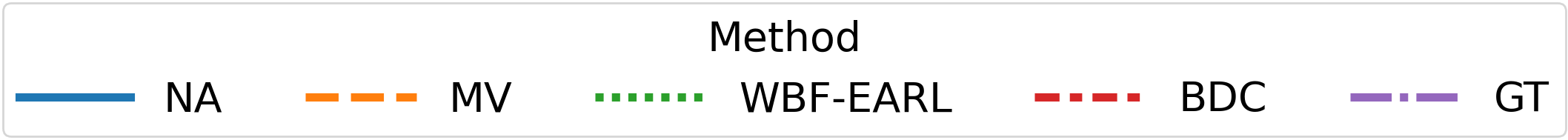}
    \end{subfigure}
    \begin{subfigure}{0.47\textwidth}
        \centering \includegraphics[trim={0 0.7cm 6.2cm 0.6cm}, clip,width=\linewidth]{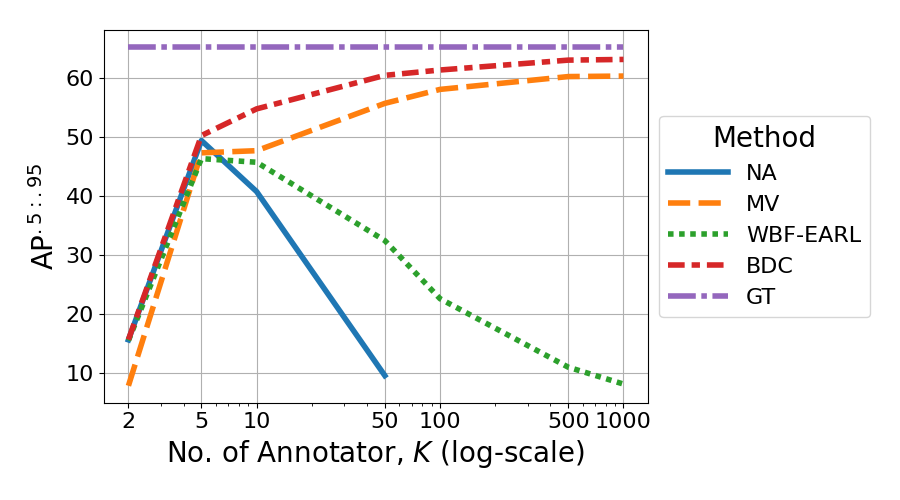}
        \caption{Scalability of BDC}\label{fig:k_ablation}
    \end{subfigure}
    \begin{subfigure}{0.47\textwidth}
        \centering \includegraphics[trim={0 0.7cm 6.2cm 0.6cm}, clip,width=\linewidth]{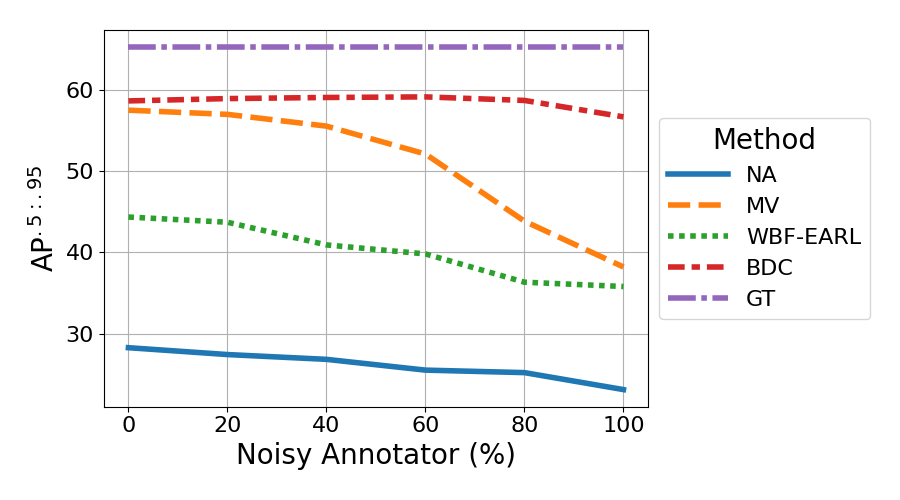}
        \caption{Robustness of BDC}\label{fig:nr_ablation}
    \end{subfigure}
    \caption{(a) Effect of the number of annotators, $K$, and (b) effect of varying percentage of noisy annotators (with $K=25$) on the test AP$^{.5:.95}$ of YOLOv7. Results for the `NA' method are unavailable when $K \ge 100$ due to GPU out-of-memory error. }
\end{figure}

We analyse the scalability and robustness of the aggregation methods using the Pascal VOC dataset. YOLOv7 is selected as the OD because it is lightweight and fast to train. \Cref{fig:k_ablation} demonstrates the effect of varying numbers of annotators, $K$, where all annotations are synthesised using annotators of average skill level. Note that training YOLOv7 with the NA method results in a GPU out-of-memory error when $K \ge 100$. When the number of annotators increases, the performance of YOLOv7 increases for both BDC and MV while the performance of both NA and WBF-EARL decreases when $K>50$. In addition, BDC has an average AP$^{.5:.95}$ improvement of 4.35 over MV.

We also investigate the effect of changing the percentage of noisy annotators while setting $K$ to 25. Here, noisy annotators are defined as annotators with a poor skill level while the other annotators are annotators with expert skill level. As illustrated in \Cref{fig:nr_ablation}, we found that AP$^{.5:.95}$ of all aggregation methods reduce when we increase the percentage of noisy annotators. However, MV drops at a faster rate compared to the proposed method when the noisy annotator percentage is over $50\%$. We observe that BDC remains robust even when 100\% of the annotators have low skill levels and the AP$^{.5:.95}$ is 18.5 higher than that obtained with MV. The consistent performance of BDC (up to 80\% noise) can be attributed to the ability of BDC to correctly capture the ability of each annotator and assign proper weights when aggregating the annotations (See Appendix~\ref{appendix:posterior_analysis} for an analysis of the posterior distributions learnt in the aggregators).

\subsection{Computational Cost} 
\label{sec:time_analysis}
\Cref{table:time_analysis} presents the wall time of each annotation aggregation method spent to pre-process the annotations and train the Faster R-CNN (FRCNN) \cite{fasterrcnn2015} model for one epoch on the VOC-MIX synthetic dataset. The wall time is computed with a desktop workstation with an i9-10900X CPU and an RTX 3090 GPU. As shown in \Cref{table:time_analysis}, MV, Crowd R-CNN and WBF-EARL pre-process the annotations before model training while NA and the proposed BDC require no pre-processing step. Crowd R-CNN has the longest training time which increases linearly with the number of annotators, $K$. On the other hand, the proposed BDC performs the annotations-predictions matching step and the updating steps of the aggregators' parameters every training epoch and therefore has a slightly longer training time compared to NA, MV and WBF-EARL. However, this increase in training time is small when compared to the much higher training cost of the OD module (618.6s OD training time out of 662.9s). Note that the model inference time is unaffected by the type of annotation aggregation method.

\begin{table}[tb]
    \caption{Annotations pre-processing time and FRCNN model training time per epoch of different annotation aggregation methods on the VOC-MIX dataset. The wall time is computed on a desktop workstation with an i9-10900X CPU and an RTX 3090 GPU.}
    \label{table:time_analysis}
    \centering
    \setlength{\tabcolsep}{5pt}
    \resizebox{.73\columnwidth}{!}{
    \begin{tabular}{l|c|c}
        \hline
        Method & Pre-processing time (s) & Training time (s) \\
        \hline
        NA & 0.0 & 537.0 \\
        MV & 19.4 & 505.4 \\
        Crowd R-CNN \cite{crowdrcnn2020} & 15.6 & 1175.5 \\
        WBF-EARL \cite{Le2023} & 4.9 & 518.5 \\
        BDC (ours) & 0.0 & 662.9 \\
        \hline
    \end{tabular}
    }
\end{table}

%% file: sec/6_conclusion.tex
\section{Conclusion}
\label{sect:conclusion}
We first presented a general annotation aggregation and object detection framework, named Bayesian Detector Combination (BDC), for learning a robust object detector from noisy crowdsourced annotations. BDC infers the consensus bounding boxes and corresponding soft labels without the need for ground truth annotations. BDC is model-agnostic and we demonstrated its superior performance by integrating it with three popular object detector backbones, namely YOLOv7, FRCNN, and EVA. Furthermore, we constructed four synthetic datasets to simulate different crowdsourcing scenarios involving annotators of various skill levels, enabling the systematic evaluation of the proposed BDC. We demonstrated the improved performance from BDC on real-world datasets and all the synthetic datasets when compared to other object detectors designed for crowdsourced annotations. We showed that BDC can scale against varying numbers of annotators and is robust against annotators with low annotation reliability.

%% file: sec/X_suppl.tex
This supplementary material provides the derivation of the updating steps of the Bounding Box Aggregator  and Class Label Aggregator in \Cref{appendix:bba,appendix:CLA}, respectively, as well as the overall training steps and pseudocode of the Bayesian Detector Combination (BDC) framework in \Cref{appendix:VB}. Moreover, \Cref{appendix:hyperparams} presents the training configuration of the experiments while \Cref{appendix:pseudocode_synthesise,appendix:visualise_networks} further discuss the synthesisation procedure to generate synthetic crowdsourced annotations. Besides, \Cref{appendix:bdc_extension} presents additional experiment results for the setting where we use the predictions of object detectors as crowdsourced annotations.
Lastly, \Cref{appendix:disaster_response,appendix:add_qualitative} show additional qualitative detection results on the disaster response dataset and the synthetic dataset, respectively, while \Cref{appendix:posterior_analysis} provides an analysis of the learnt posterior distributions of BDC on the COCO-MIX dataset.

\section{Bounding Box Aggregator}
\label{appendix:bba}

We present in this section the derivation of the equations (Equations \eqref{eq:mu_update_main}-\eqref{eq:beta_update_main}) to update the posterior distribution of the error terms introduced in \Cref{eq:box_error}. The derived error posterior mean and covariance are used to aggregate the bounding boxes (via \Cref{eq:correct_box} and \Cref{eq:aggregate_box}). These equations are used in the training pipeline as described in Lines 7, 8, and 13 of Algorithm~\ref{alg:training} in \Cref{appendix:VB}.

\paragraph{\bf Problem Setup.}
Following the notations in Section~\ref{sect:method}, for brevity, notations in this section are simplified to consider a single training sample. 
Let $(y_m, \hat{y}_m^*)_{m=1}^M$ be the pairs of annotation-prediction matches obtained from the Annotations-Predictions Matcher in \Cref{sec:matcher}, where $M$ is the number of annotation-prediction matches. $y_m$ contains the bounding box coordinates $b_m \in \mathbb{R}^4$ and their annotator ID $k_m \in \{1,...,K\}$, while $\hat{y}_m^*$ contains the matched bounding box prediction $\hat{b}_m^* \in \mathbb{R}^4$ from the object detector (OD) module.

Popular object detectors, such as YOLO \cite{yolo} and DETR \cite{detr2020}, do not predict the bounding box coordinates in the form of probability distributions. Instead, they predict four values representing the x and y coordinates of the box centre along with the width and height of the box. Hence, we cannot directly perform Bayesian inference for the bounding box prediction. To address this, we model the four error terms, \ie the translation error in the x-axis, the translation error in the y-axis, the scaling error in the x-axis and the scaling error in the y-axis, of each annotator by multivariate Gaussian distributions:
\begin{equation}
    p(\epsilon_m | k_{m}=k, \bm{\mu}, \bm{\sigma}) = \mathcal{N}({\mu}^k, {\sigma}^k)\,\,,
\end{equation}
where $\epsilon_m \in \mathbb{R}^4$ is a vector containing the four error terms while $\bm{\mu}$ and $\bm{\sigma}$ are a collection of Gaussian parameters with ${\mu}^k$ and ${\sigma}^k$ representing the mean and covariance of $\epsilon_m$ for the $k$-th annotator, respectively. 
The four error terms, which represent the differences between the matched prediction and annotated bounding box from each annotator, are calculated using \Cref{eq:box_error} in \Cref{sec:BBA}.

In this section, the goal is to derive the posterior distribution of these error terms shown in Equations \eqref{eq:mu_update_main}-\eqref{eq:beta_update_main}.

\paragraph{\bf Initialisation.}
We impose a Gaussian-Gamma \cite{gaussgamma} prior with hyperparameters, $\mu^k_0$, $\upsilon^k_0$ and $\beta^k_0$, for the Gaussian distributions:
\begin{align}
    \mathcal{NG}({\mu}^k, {\sigma}^k| \mu^k_0, \upsilon^k_0, \beta^k_0) &\stackrel{\text{def}}{=} \mathcal{N}(\mu^k|\mu^k_0, \sigma^k) \mathcal{G}((\sigma^k)^{-1}|\upsilon^k_0, \beta^k_0) \nonumber \\
        &= \frac{1}{(2 \pi)^2} \frac{1}{\sqrt{|\sigma^k|}} \exp \left( -\frac{1}{2} (\mu^k - \mu^k_0)^\top(\sigma^k)^{-1}(\mu^k - \mu^k_0) \right) \nonumber \\
        &\quad \times \frac{|\beta^k_0| ^ {\upsilon^k_0}}{\Gamma(\upsilon^k_0)} \frac{1}{|\sigma^k|^{\upsilon^k_0 - \frac{5}{2}}} \exp \left( -tr(\beta^k_0(\sigma^k)^{-1}) \right)\,\,. 
    \label{eq:gaussgamma}
\end{align}
Here, $\mathcal{NG}$ denotes the Gaussian-Gamma distribution, $\mathcal{N}$ represents the Gaussian distribution, $\mathcal{G}$ represents the Gamma distribution, while $|\cdot|$ denotes the determinant of a matrix, $tr(\cdot)$ is the trace of a matrix, and $\Gamma$ denotes the Gamma function \cite{gammafunc}.

\paragraph{\bf Evidence Lower Bound.}
Let $\mathbf{E}$ represent the set of all the bounding box errors $(\epsilon_m, k_m)$, while $\mathbf{M}_0$, $\mathbf{U}_0$ and $\mathbf{B}_0$ are the Gaussian-Gamma prior hyperparameters $\mu^k_0$, $\upsilon^k_0$ and $\beta^k_0$, respectively. The log marginal likelihood of the observed bounding box errors $\mathbf{E}$ can be computed as:

\begin{align}
    \ln\, p(\mathbf{E} | \mathbf{M}_0, \mathbf{U}_0, \mathbf{B}_0) &= \underbrace{\int q(\bm{\mu}, \bm{\sigma}) \ln \frac{p(\mathbf{E}, \bm{\mu}, \bm{\sigma} | \mathbf{M}_0, \mathbf{U}_0, \mathbf{B}_0)}{q(\bm{\mu}, \bm{\sigma})} \diff\bm{\mu} \diff\bm{\sigma} \,}_{\mathcal{L}_b(q(\bm{\mu}, \bm{\sigma}))} \label{eq:elbo_box} \\
       & \quad + \underbrace{\int q(\bm{\mu}, \bm{\sigma}) \ln \frac{q(\bm{\mu}, \bm{\sigma})}{p(\bm{\mu}, \bm{\sigma} | \mathbf{E}, \mathbf{M}_0, \mathbf{U}_0, \mathbf{B}_0)} \diff\bm{\mu} \diff\bm{\sigma} \,}_{KL(q(\bm{\mu}, \bm{\sigma}) || p(\bm{\mu}, \bm{\sigma}))} \,\,,
\end{align}
where $q(\bm{\mu}, \bm{\sigma})$ is an arbitrary distribution of $\bm{\mu}$ and $\bm{\sigma}$ while $KL$ is the Kullback-Leibler (KL) divergence. The joint likelihood in \Cref{eq:elbo_box} can be rewritten as:
\begin{equation}
    p(\mathbf{E},\bm{\mu}, \bm{\sigma}| \mathbf{M}_0, \mathbf{U}_0, \mathbf{B}_0) = p(\mathbf{E}|\bm{\mu}, \bm{\sigma}) p(\bm{\mu}, \bm{\sigma}|\mathbf{M}_0, \mathbf{U}_0, \mathbf{B}_0) \,\,.
    \label{eq:box_joint}
\end{equation}

Since the KL divergence is always non-negative, $\mathcal{L}_b(q(\bm{\mu}, \bm{\sigma}))$ is the lower bound of the marginal likelihood, which we can rewrite as:
\begin{align}
    \mathcal{L}_b(q(\bm{\mu}, \bm{\sigma})) &= - \int q(\bm{\mu}, \bm{\sigma}) \ln \frac{q(\bm{\mu}, \bm{\sigma})}{p(\mathbf{E}, \bm{\mu}, \bm{\sigma} | \mathbf{M}_0, \mathbf{U}_0, \mathbf{B}_0)} \diff\bm{\mu} \diff\bm{\sigma} \nonumber \\
    &= - KL(q(\bm{\mu}, \bm{\sigma}) || p(\mathbf{E}, \bm{\mu}, \bm{\sigma} | \mathbf{M}_0, \mathbf{U}_0, \mathbf{B}_0)) \,\,.
    \label{eq:fact_elbo_box}
\end{align}

\paragraph{\bf Optimising $\mathcal{L}_b$.}
We seek to maximise the lower bound (\Cref{eq:fact_elbo_box}), which is equivalent to minimising the KL divergence between $q(\bm{\mu}, \bm{\sigma})$ and $p(\mathbf{E}, \bm{\mu}, \bm{\sigma} | \mathbf{M}_0, \allowbreak \mathbf{U}_0, \mathbf{B}_0)$. The KL divergence is minimised when the two distributions are equivalent:
\begin{align}
    \ln\ &q(\bm{\mu}, \bm{\sigma}) = \ln p(\mathbf{E},\bm{\mu}, \bm{\sigma}| \mathbf{M}_0, \mathbf{U}_0, \mathbf{B}_0) \nonumber \\
    &\;\;\,\stackequal{C}{\cref{eq:box_joint}, \cref{eq:gaussgamma}} \sum_{k=1}^K \ln \frac{1}{\sqrt{|\sigma^k|}} \frac{1}{|\sigma^k|^{\upsilon^k_0 + \frac{M^k}{2} - \frac{5}{2}}} \exp \left( -tr(\beta^k_0(\sigma^k)^{-1}) \right) \nonumber \\ &\qquad\qquad\quad \exp \Bigg( - \frac{1}{2} \Big[(\mu^k - \mu^k_0)^\top(\sigma^k)^{-1}(\mu^k - \mu^k_0) \nonumber \\
    &\qquad\qquad\qquad + \sum_{m=1}^{M} \mathbbm{1}[k_m=k](\epsilon_m - \mu^k)^\top(\sigma^k)^{-1}(\epsilon_m - \mu^k) \Big] \Bigg) \,\,,
    \label{eq:q_mu_sigma}
\end{align}
where $\mathbbm{1}[\cdot]$ is the indicator function, $M^k := \sum_{m=1}^M \mathbbm{1}[k_m=k]$ is the number of annotation-prediction matches associated with the $k$-th annotator, and $\stackrel{C}{=}$ denotes equality up to an additive constant that is independent of the function parameters.

Let us define the mean of all $\epsilon_m$ computed using \Cref{eq:box_error} for the $k$-th annotator as follow:
\begin{equation}
    \bar{\epsilon}^k = \frac{1}{M^k}\sum_{m=1}^M \mathbbm{1}[k_m=k]\epsilon_m \,\,.
\end{equation}
We can then rewrite one term in the exponent of \Cref{eq:q_mu_sigma} as follows:
{\small
\begin{align}
    \sum_{m=1}^{M} &\mathbbm{1}[k_m=k](\epsilon_m - \mu^k)^\top(\sigma^k)^{-1}(\epsilon_m - \mu^k) \nonumber \\ &= \sum_{m=1}^{M} \mathbbm{1}[k_m=k][(\epsilon_m - \bar{\epsilon}^k) - (\mu^k - \bar{\epsilon}^k)]^\top (\sigma^k)^{-1}[(\epsilon_m - \bar{\epsilon}^k) - (\mu^k - \bar{\epsilon}^k)] \\
    &= \sum_{m=1}^{M} \mathbbm{1}[k_m=k](\epsilon_m - \bar{\epsilon}^k)^\top(\sigma^k)^{-1}(\epsilon_m - \bar{\epsilon}^k) 
     + M^k(\mu^k - \bar{\epsilon}^k)^\top(\sigma^k)^{-1}(\mu^k - \bar{\epsilon}^k)\,\,.
    \label{eq:sum_squares}
\end{align}
}
It can also be shown that:
\begin{align}
    (\mu^k - \mu^k_0)&^\top(\sigma^k)^{-1}(\mu^k - \mu^k_0) + M^k(\mu^k - \bar{\epsilon}^k)^\top(\sigma^k)^{-1}(\mu^k - \bar{\epsilon}^k) \nonumber \\
    &= (M^k + 1)\left(\mu^k - \frac{\mu^k_0 + M^k\bar{\epsilon}^k}{M^k + 1} \right)^\top(\sigma^k)^{-1}\left(\mu^k - \frac{\mu^k_0 + M^k\bar{\epsilon}^k}{M^k + 1} \right) \nonumber \\ &\quad + \frac{M^k(\bar{\epsilon}^k - \mu^k_0)^\top(\sigma^k)^{-1}(\bar{\epsilon}^k - \mu^k_0)}{M^k + 1}\,\,.\label{eq:gauss_exp_term}
\end{align}

By plugging \Cref{eq:sum_squares} and \Cref{eq:gauss_exp_term} into \Cref{eq:q_mu_sigma}, we obtain
{\small
\begin{align}
    \ln\ q(\bm{\mu}, \bm{\sigma}) &\stackrel{C}{=} \sum_{k=1}^K \ln \frac{1}{\sqrt{|\sigma^k}|} \frac{1}{|\sigma^k|^{\upsilon^k_0 + \frac{M^k}{2} - \frac{5}{2}}} \exp \left( -tr(\beta^k_0(\sigma^k)^{-1}) \right) \nonumber \\
    &\qquad\ \exp \Bigg( - \frac{1}{2} \Bigg[(M^k + 1) \left(\mu^k - \frac{\mu^k_0 + M^k\bar{\epsilon}^k}{M^k + 1} \right)^\top(\sigma^k)^{-1}\left(\mu^k - \frac{\mu^k_0 + M^k\bar{\epsilon}^k}{M^k + 1} \right) \nonumber \\
    &\qquad \quad + \frac{M^k(\bar{\epsilon}^k - \mu^k_0)^\top(\sigma^k)^{-1}(\bar{\epsilon}^k - \mu^k_0)}{M^k + 1} \nonumber \\
    &\qquad \quad + \sum_{m=1}^{M} \mathbbm{1}[k_m=k](\epsilon_m - \bar{\epsilon}^k)^\top(\sigma^k)^{-1}(\epsilon_m - \bar{\epsilon}^k) \Bigg] \Bigg) \nonumber \\
    &\stackrel{C}{=} \sum_{k=1}^K \ln \frac{1}{\sqrt{|\sigma^k}|} \exp\Bigg( - \frac{1}{2} (M^k + 1) \left(\mu^k - \frac{\mu^k_0 + M^k\bar{\epsilon}^k}{M^k + 1} \right)^\top(\sigma^k)^{-1} \nonumber \\ &\qquad\qquad\qquad\qquad \left(\mu^k - \frac{\mu^k_0 + M^k\bar{\epsilon}^k}{M^k + 1} \right) \Bigg) \nonumber\\
    &\qquad \times \frac{1}{|\sigma^k|^{\upsilon^k_0 + \frac{M^k}{2} - \frac{5}{2}}} \exp \Bigg( -tr\Bigg(\bigg[ \beta^k_0 + \frac{M^k(\bar{\epsilon}^k - \mu^k_0)(\bar{\epsilon}^k - \mu^k_0)^\top}{2(M^k + 1)} \nonumber \\
    &\qquad\quad + \frac{1}{2}\sum_{m=1}^{M} \mathbbm{1}[k_m=k](\epsilon_m - \bar{\epsilon}^k)(\epsilon_m - \bar{\epsilon}^k)^\top \bigg](\sigma^k)^{-1} \Bigg) \Bigg) \,\,.
\end{align}
}
We can therefore conclude that $q(\bm{\mu}, \bm{\sigma})$ is the product of the Gaussian-Gamma distributions (\Cref{eq:gaussgamma}) \cite{normalconj}:
\begin{equation}
    q(\bm{\mu}, \bm{\sigma}) = \prod_{k=1}^K \mathcal{NG}({\mu}^k, {\sigma}^k| \Tilde{\mu}^k, \Tilde{\upsilon}^k, \Tilde{\beta}^k)\,\,,
\end{equation}
where
\begin{equation}
\label{eq:mu_update}
    \Tilde{\mu}^k = \frac{\mu^k_0 + M^k \bar{\epsilon}^k}{M^k + 1}\,\,,
\end{equation}
\begin{equation}
\label{eq:upsilon_update}
    \Tilde{\upsilon}^k = \upsilon^k_0 + \frac{M^k}{2}\,\,,
\end{equation}
\begin{equation}
\label{eq:beta_update}
    \Tilde{\beta}^k = \beta^k_0 + \frac{M^k(\bar{\epsilon}^k - \mu^k_0)(\bar{\epsilon}^k - \mu^k_0)^\top}{2(M^k + 1)} + \frac{1}{2}\sum_{m=1}^{M} \mathbbm{1}[k_m=k](\epsilon_m - \bar{\epsilon}^k)(\epsilon_m - \bar{\epsilon}^k)^\top\,\,.
\end{equation}

Finally, we obtain the expected mean and covariance matrix of the updated Gaussian-Gamma posterior as $\mu^k = \Tilde{\mu}^k$ and $\sigma^k = \Tilde{\beta}^k / \Tilde{\upsilon}^k$, respectively, for the aggregation of the bounding boxes (via \Cref{eq:correct_box} and \Cref{eq:aggregate_box}).

\newpage

\section{Class Label Aggregator}
\label{appendix:CLA}

We present in this section the derivation of Equation \eqref{eq:rho} to update the aggregated class labels to maximise the evidence lower bound (ELBO) of the crowdsourced labels. We also provide the derivation to update each annotator's confusion matrix to maximise the ELBO via \Cref{eq:alpha_update}. These updates correspond to Lines 9 and 14, respectively, in Algorithm~\ref{alg:training} of the training pipeline described in \Cref{appendix:VB} and supplement the material introduced in \Cref{sec:CLA}.

\paragraph{\bf Problem Setup.}

As in \Cref{sect:method}, the notations in this section are simplified to consider a single training sample for brevity.
The derivation steps follow those of the Bayesian classifier combination in Kim and Ghahramani~\cite{BCC2012} and Isupova \etal~\cite{bccnet2018}.

Let $\{(c_m, k_m)\}_{m=1}^{M_n}$ be the set of tuples consisting of the annotated class labels $c_m \in \{1,...,J\}$ and their annotator ID $k_m \in \{1,...,K\}$ that are matched to the $n$-th prediction $\hat{y}_n$ obtained from the Annotations-Predictions Matcher in \Cref{sec:matcher}. $\hat{y}_n$ contains the predicted class probabilities $\hat{p}_n \in [0, 1]^J$, and $M_n$ denotes the number of annotations matched to the $n$-th prediction.

In the CLA, we model the annotated class label from each annotator as a multinomial distribution conditioning on the true label of the object: 
\begin{equation}
    p(c_m | k_m = k, t_m = j, \bm{\pi}) = {\pi}^k_{j, c_m} \,\,,
\end{equation}
where $\bm{\pi}$ is a collection of confusion matrices, ${\pi}^k \in \mathbb{R}^{J \times J}$ is the confusion matrix of the $k$-th annotator and $t_m$ is the ground truth class label. Each row of $\pi_j^k$ represents the event probabilities of a multinomial distribution for the annotated class label from the $k$-th annotator when $t_m=j$. The probability of the true class label of the $n$-th object, $p(t_n|x, \theta_O)$ is modelled by the OD module with a softmax output layer.

In this section, the goal is to learn the set of distributions for each confusion matrix in $\bm{\pi}$ (Equations \eqref{eq:q_pi_dir} and \eqref{eq:alpha_update}) and the distribution of the aggregated (soft) ground truth label for the $n$-th prediction $\rho_{n,j} := q(t_n=j)$ (\Cref{eq:rho}).

\paragraph{\bf Initialisation.}
We impose a Dirichlet prior with hyperparameters ${\alpha}^k_{0j} = \{\alpha^k_{0j,1},\allowbreak ..., \alpha^k_{0j,J}\}$ such that $\pi^k_j | \alpha^k_{0j} \sim Dir(\alpha^k_{0j})$.

\paragraph{\bf Evidence Lower Bound.}
Let $\mathbf{c}$ capture all annotated class labels $(c_m, k_m)$, $\mathbf{t}$ are all the true labels $t_n$, $\mathbf{A}_0$ are all the prior hyperparameters $\alpha_{0j}^k$ for the Dirichlet priors. The log marginal likelihood of the observed class labels $\mathbf{c}$ can be described as:

\begin{align}
    \ln\, p(\mathbf{c} | x, \mathbf{A}_0, \theta_O)
       &=\underbrace{\int q(\mathbf{t}, \bm{\pi}) \ln \frac{p(\mathbf{c},\mathbf{t}, \bm{\pi} | x, \mathbf{A}_0, \theta_O)}{q(\mathbf{t}, \bm{\pi})} \diff\mathbf{t}  \diff\bm{\pi}}_{\mathcal{L}_c(q, \theta_O)}\\
       &\quad + \underbrace{\int q(\mathbf{t}, \bm{\pi}) \ln \frac{q(\mathbf{t}, \bm{\pi})}{p(\mathbf{t}, \bm{\pi} | \mathbf{c}, x, \mathbf{A}_0, \theta_O)} \diff\mathbf{t}  \diff\bm{\pi}}_{KL( q(\mathbf{t}, \bm{\pi})|| p(\mathbf{t}, \bm{\pi}))} \,\,,
\end{align}
where $q(\mathbf{t}, \bm{\pi})$ is an arbitrary distribution of $\mathbf{t}$ and $\bm{\pi}$ and $\mathcal{L}_c(q, \theta_O)$ is the lower bound of the marginal likelihood. 

We employ mean-field variational inference \cite{wainwright2008graphical} and restrict our search within factorised distributions $q(\mathbf{t}, \bm{\pi}) = q(\mathbf{t})q(\bm{\pi})$:
\begin{multline}
    \mathcal{L}_c(q, \theta_O) = \int q(\mathbf{t})q(\bm{\pi}) \ln p(\mathbf{c},\mathbf{t}, \bm{\pi} | x, \mathbf{A}_0, \theta_O) \diff\mathbf{t}  \diff\bm{\pi} \\
     - \int q(\mathbf{t}) \ln q(\mathbf{t}) \diff\mathbf{t} - \int q(\bm{\pi}) \ln q(\bm{\pi}) \diff\bm{\pi} \,.
     \label{eq:fact_elbo_class}
\end{multline}
The lower bound, $\mathcal{L}_c(q, \theta_O)$, depends on the joint likelihood which can be factorised as:
\begin{equation}
    p(\mathbf{c},\mathbf{t}, \bm{\pi} | x, \mathbf{A}_0, \theta_O) = p(\mathbf{c}|\mathbf{t}, \bm{\pi}) p(\mathbf{t}| x, \theta_O) p(\bm{\pi}|\mathbf{A}_0)\,\,.
    \label{eq:joint}
\end{equation}

\paragraph{\bf Optimising $\mathcal{L}_c$ with respect to $q(\bm{\pi})$.}
First, re-writing the lower bound as a function of $q(\bm{\pi})$ with:
\begin{align}
    \ln \Tilde{p}_{\bm{\pi}} (\mathbf{c},\mathbf{t}, \bm{\pi}) :&= \int q(\mathbf{t}) \ln p(\mathbf{c},\mathbf{t}, \bm{\pi} | x, \mathbf{A}_0, \theta_O) \diff\mathbf{t} \nonumber \\
    &\stackrel{C}{=} \mathbb{E}_{q(\mathbf{t})} \ln p(\mathbf{c},\mathbf{t}, \bm{\pi} | x, \mathbf{A}_0, \theta_O) \,\,,
\end{align} 
gives:
\begin{align}
    \mathcal{L}_c(q(\bm{\pi})) &\stackrel{C}{=} -\int q(\bm{\pi}) \ln \frac{q(\bm{\pi})}{\Tilde{p}_{\bm{\pi}} (\mathbf{c},\mathbf{t}, \bm{\pi})}  \diff\bm{\pi} \nonumber \\
    &\stackrel{C}{=} -KL(q(\bm{\pi}) || \Tilde{p}_{\bm{\pi}} (\mathbf{c},\mathbf{t}, \bm{\pi})) \,\,.
    \label{eq:elbo_q_pi}
\end{align}
In other words, to maximise $\mathcal{L}_c(q(\bm{\pi}))$, we need to minimise the KL divergence between $q(\bm{\pi})$ and $ \Tilde{p}_{\bm{\pi}} (\mathbf{c},\mathbf{t}, \bm{\pi})$ which is minimised when the two distributions are equal:
{\small
\begin{align}
    \ln\ &q(\bm{\pi}) \stackrel{C}{=}\ \mathbb{E}_{q(\mathbf{t})} \ln p(\mathbf{c},\mathbf{t}, \bm{\pi} | x, \mathbf{A}_0, \theta_O) \nonumber \\
    &\quad \stackequal{C}{\cref{eq:joint}} \mathbb{E}_{q(\mathbf{t})} \ln p(\mathbf{c}|\mathbf{t}, \bm{\pi}) p(\bm{\pi}|\mathbf{A}_0) \label{eq:q_pi} \\
        &\qquad \stackrel{C}{=}\sum_{k=1}^K \sum_{j=1}^J \sum_{l=1}^J \ln \pi^{k}_{j,l} \Bigg(\alpha^k_{0j,l} + \sum_{n=1}^H \sum_{(c,k')\in \Tilde{\kappa}_n} \mathbbm{1}[c=l]\mathbbm{1}[k'=k] q(t_n=j) -1 \Bigg) \,.
\end{align}
}
$H$ indicates the number of predictions matched to at least one annotation, $\Tilde{\kappa}_n$ is the set of tuples consisting of the annotated class labels and their annotator ID that are matched to the $n$-th prediction while $\mathbbm{1}[\cdot]$ is the indicator function. We observe that $q(\bm{\pi})$ is the product of the Dirichlet distributions:
\begin{equation}
    q(\bm{\pi}) = \prod_{k=1}^K \prod_{j=1}^J Dir(\pi_j^k|\Tilde{\alpha}_j^k)\,\,,
    \label{eq:q_pi_dir}
\end{equation}
where the Dirichlet parameters are updated with:
\begin{equation}
\label{eq:alpha_update}
    \Tilde{\alpha}^k_{j,l} = \alpha^k_{0j,l} + \sum_{n=1}^H \sum_{(c,k') \in \Tilde{\kappa}_n} \mathbbm{1}[c = l]\mathbbm{1}[k' = k]\,q(t_n=j)\,\,.
\end{equation}

\paragraph{\bf Optimising $\mathcal{L}_c$ with respect to $q(\mathbf{t})$.}
Next, we rewrite the lower bound as a function of $q(\mathbf{t})$ and follow similar steps of \Cref{eq:elbo_q_pi,eq:q_pi} to obtain the updating formula for the distribution $q(\mathbf{t})$:
\begin{align}
    \ln &\ q(\mathbf{t}) \stackrel{C}{=}\ \mathbb{E}_{q(\bm{\pi})} \ln p(\mathbf{c},\mathbf{t}, \bm{\pi} | x, \mathbf{A}_0, \theta_O) \nonumber \\
    &\quad \stackequal{C}{\cref{eq:joint}} \mathbb{E}_{q(\bm{\pi})} \ln p(\mathbf{c}|\mathbf{t}, \bm{\pi}) p(\mathbf{t}| x, \theta_O) \nonumber \\
        &\qquad \stackrel{C}{=}\sum_{n=1}^H \sum_{j=1}^J \mathbbm{1}[t_n=j] \left(\sum_{(c,k) \in \Tilde{\kappa}_n}\mathbb{E}_{\pi^k_j}[\ln \pi^{k}_{j, c}] + \ln p(t_n=j|x, \theta_O) \right) \,\,. 
    \label{eq:q_t}
\end{align}
We derived that $q(\pi^k_j)$ are the Dirichlet distributions (\Cref{eq:q_pi_dir}). Therefore, the expected value for each annotator is given by:
\begin{equation}
    \mathbb{E}_{\pi^{k}_j}\ln \pi^{k}_{j, c} = \Psi(\Tilde{\alpha}^{k}_{j,c}) - \Psi\left(\sum^J_{l=1} \Tilde{\alpha}^{k}_{j,l}\right) \,\,,
\end{equation}
where $\Psi$ denotes the digamma function \cite{digamma}.

Using the definition of categorical distribution (\ie if $p(a)$ is a categorical distribution, then its logarithm is given as $\ln p(a) = \sum_i \mathbbm{1}[a=i]\ln p(a=i)$), we conclude that $q(t)$ is the product of categorical distributions from \Cref{eq:q_t}:
\begin{equation*}
    q(t) = \prod_{n=1}^H q(t_n) \,\,,
\end{equation*}
such that the probability of the true label for the $n$-th object, $\rho_{n,j} := q(t_n=j)$, is:
\begin{equation}
    \rho_{n, j} = \exp \left(\ln \hat{p}_{n, j} + \sum_{(c,k) \in \Tilde{\kappa}_n} \mathbb{E}_{\pi^k_j}\ln \pi^k_{j, c}\right)\,\,,
    \label{eq:rho_appendix}
\end{equation}
where $\hat{p}_{n,j} := p(t_n=j|x,\theta_O)$ is the OD’s class-$j$ probability of the $n$-th prediction.

\newpage
\section{Training Pipeline and Pseudocode}
\label{appendix:VB}

\Cref{alg:training} summarises the overall training steps of the BDC framework.%

\begin{algorithm}
    \small
    \caption{Pseudocode for the Training Steps of BDC.}
    \label{alg:training}
    \begin{algorithmic}[1]
        \Require{Set of input images, $x=\left\{x^i\in \mathbb{R}^{I_w \times I_h}\right\}_{i=1}^N$, where $I_w$ and $I_h$ are the width and height of the image, respectively, while $N$ is the number of training images; Set of crowdsourced annotations, $y=\left\{y^i = \big\{y^i_m = (b^i_m, c^i_m, k^i_m)\right\}_{m=1}^{M^i}\big\}_{i=1}^N$, where $y^i_m$ is a tuple of the annotated bounding box $b^i_m \in \mathbb{R}^4$, the annotated class label $c^i_m \in \{1,...,J\}$, and the annotator ID $k^i_m \in \{1,...,K\}$ of the $m$-th annotation in the $i$-th image, while $M^i$ is the number of crowdsourced annotations in the $i$-th image; The number of training epochs, $T \in \mathbb{Z}^+$; The object detector (OD) module, $\textit{OD}_{\theta_O}(\cdot):\mathbb{R}^{I_w \times I_h} \to O^P$, where $\theta_O$ is the initialised OD's parameters, $O=\mathbb{R}^4 \times [0,1]^J$ is the predicted box coordinates and class probabilities while $P$ is the number of predictions; The bounding box aggregator (BBA) with the initialised parameters, $\bm{\mu}_0$, $\bm{\upsilon}_0$ and $\bm{\beta}_0$; The class label aggregator (CLA) with the initialised parameters, $\bm{\alpha}_0$.}
            \State $t \gets 1$;
            \While{all parameters are not converged and $t \leq T$}
                \State Initialise $\Hat{y} \gets \emptyset$, $\mathfrak{b} \gets \emptyset$, $\rho \gets \emptyset$;
                \For{$i=1$ to $N$}
                    \State $\Hat{y}^i \gets$ $\textit{OD}_{\theta_O}(x^i)$;\Comment{Forward pass of the OD to generate predictions.}
                    \State Match each crowdsourced annotation to one of the predictions by minimising the matching cost (\Cref{eq:matching_cost});
                    \State Adjust the annotated bounding boxes (\Cref{eq:correct_box}) using the posterior mean of the error terms in BBA;
                    \State $\mathfrak{b}^i \gets$ Aggregate the bounding boxes (\Cref{eq:aggregate_box}) using the posterior precision of the error terms in BBA;
                    \State $\rho^i \gets$ Obtain the aggregated class probabilities (\Cref{eq:rho}) using the current value of the CLA's parameters;
                    \State $\Hat{y} := \Hat{y} \cup \Hat{y}^i$;\, $\mathfrak{b} := \mathfrak{b} \cup \mathfrak{b}^i$;\, $\rho := \rho \cup \rho^i$;
                \EndFor
                \State Update $\theta_O$ by minimising the OD loss function (\Cref{eq:od_loss}) via the gradient descent method;
                \State Update $q(\bm{\mu}, \bm{\sigma})$ according to Equations \eqref{eq:mu_update}-\eqref{eq:beta_update} to maximise $\mathcal{L}_b$;
                \State Update $q(\bm{\pi})$ according to \Cref{eq:alpha_update} to maximise $\mathcal{L}_c$;
            \EndWhile
    \end{algorithmic}
\end{algorithm}

\newpage

\section{Training Configuration}
\label{appendix:hyperparams}
\Cref{table:hyperparams} shows the training settings for each object detection algorithm. 
Code is implemented with PyTorch 1.13 \cite{pytorch} and runs on RTX Titan and 3090 GPUs. 

\begin{table}[hb]
    \setlength{\tabcolsep}{5pt}
    \caption{Training settings and configuration for YOLOv7 \cite{wang2022yolov7}, Faster R-CNN (FRCNN) \cite{fasterrcnn2015} and EVA \cite{eva2023} object detectors.}
    \label{table:hyperparams}
    \centering
        \begin{tabular}{l|ccc}
            \hline
            Setting & YOLOv7 \cite{wang2022yolov7} & FRCNN \cite{fasterrcnn2015} & EVA \cite{eva2023} \\
            \hline
            \hline
            Pre-training data & \multicolumn{2}{c}{ImageNet \cite{imagenet}} &  Merged-38M \cite{eva2023} \\
            Image size & 640 & 800 & 1,024 \\
            Optimiser & \multicolumn{2}{c}{SGD w/ momentum} & AdamW \cite{adamw} \\
            LR & \multicolumn{2}{c}{0.001} & $5e^{-5}$\\
            LR scheduler & \multicolumn{2}{c}{Cosine decay \cite{cosdecay}} & - \\
            Weight decay & \multicolumn{2}{c}{0.0005} & 0.1\\
            Batch size & 32 & 8 & 8\\
            Epochs (VOC/COCO) & 100/20 & 50/15 & 25/10\\
            \hline
        \end{tabular}

\end{table}

\newpage
\section{Results on the Disaster Response Dataset}
\label{appendix:disaster_response}
This section presents results on the disaster response dataset~\cite{bccnet2018}. As discussed in the main paper, there is no ground truth for this dataset. Therefore, we present only the qualitative performance of algorithms. It is worth noting that in~\cite{bccnet2018} the dataset was adjusted to be used for image classification. However, crowd members provided bounding boxes for buildings and labelled damages on them. Hence, these labels can be used for object detection.

\begin{figure}[ht]
    \captionsetup[subfigure]{format=hang}
    \begin{minipage}[c]{0.23\textwidth}
        (a) Before Disaster
    \end{minipage}
    \begin{minipage}[c]{0.25\textwidth}
        \centering \includegraphics[width=\linewidth,trim={0 0 0 5cm},clip]{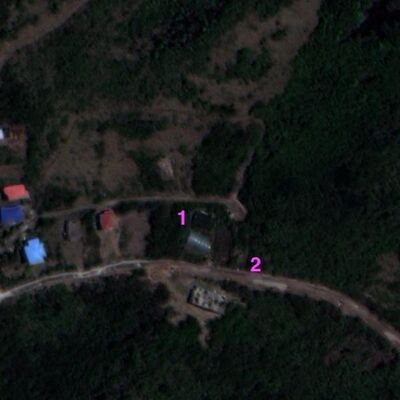}
    \end{minipage}
    \begin{minipage}[c]{0.25\textwidth}
        \centering \includegraphics[width=\linewidth,trim={0 0 0 5cm},clip]{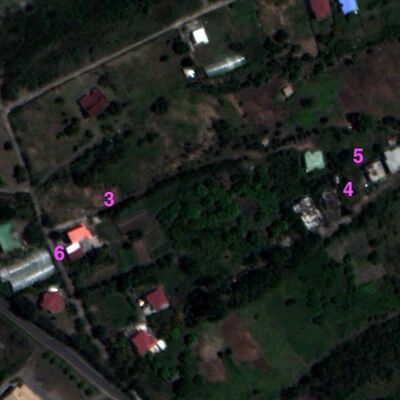}
    \end{minipage}
    \begin{minipage}[c]{0.25\textwidth}
        \centering \includegraphics[width=\linewidth,trim={0 0 0 5cm},clip]{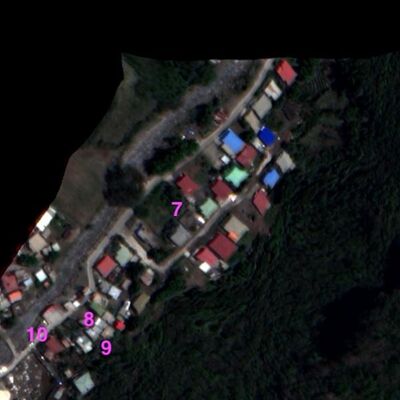}
    \end{minipage}
    \smallskip
    
    \begin{minipage}[c]{0.23\textwidth}
        (b) After Disaster
    \end{minipage}
    \begin{minipage}[c]{0.25\textwidth}
        \centering \includegraphics[width=\linewidth,trim={0 0 0 5cm},clip]{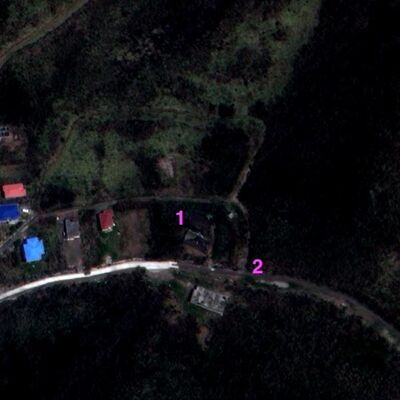}
    \end{minipage}
    \begin{minipage}[c]{0.25\textwidth}
        \centering \includegraphics[width=\linewidth,trim={0 0 0 5cm},clip]{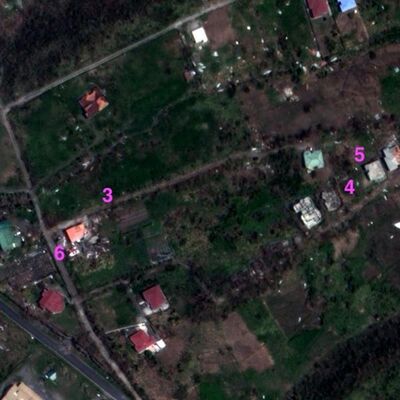}
    \end{minipage}
    \begin{minipage}[c]{0.25\textwidth}
        \centering \includegraphics[width=\linewidth,trim={0 0 0 5cm},clip]{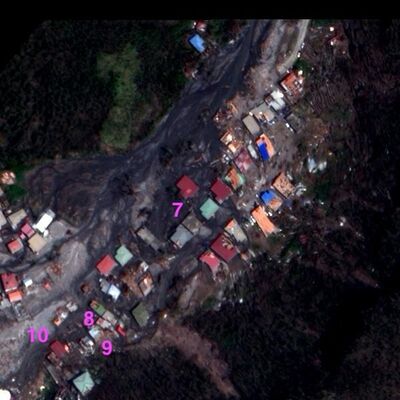}
    \end{minipage}
	\smallskip
	
    \begin{minipage}[c]{0.23\textwidth}
        (c) NA
    \end{minipage}
    \begin{minipage}[c]{0.25\textwidth}
        \centering \includegraphics[width=\linewidth,trim={0 0 0 5cm},clip]{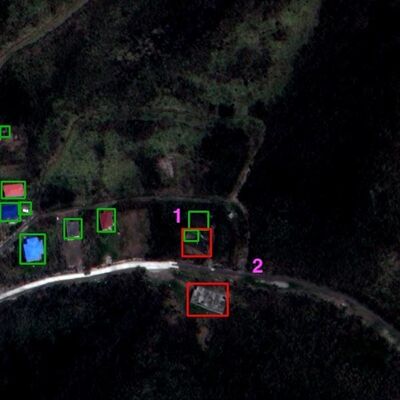}
    \end{minipage}
    \begin{minipage}[c]{0.25\textwidth}
        \centering \includegraphics[width=\linewidth,trim={0 0 0 5cm},clip]{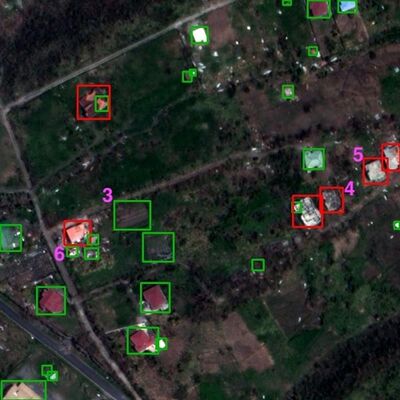}
    \end{minipage}
    \begin{minipage}[c]{0.25\textwidth}
        \centering \includegraphics[width=\linewidth,trim={0 0 0 5cm},clip]{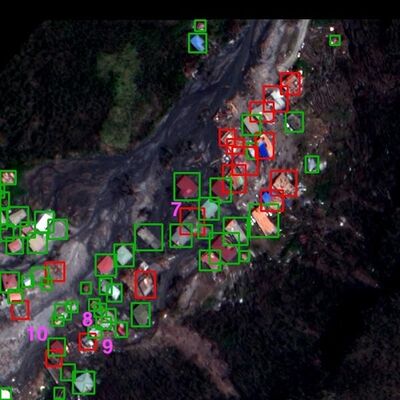}
    \end{minipage}
	\smallskip
	
    \begin{minipage}[c]{0.23\textwidth}
        (d) MV
    \end{minipage}
    \begin{minipage}[c]{0.25\textwidth}
        \centering \includegraphics[width=\linewidth,trim={0 0 0 5cm},clip]{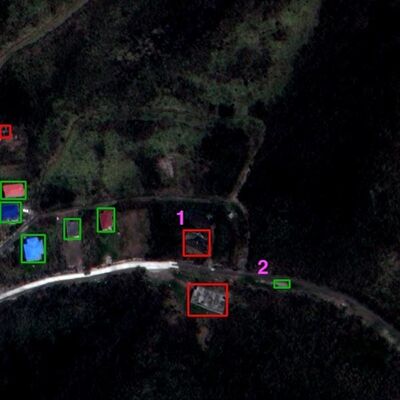}
    \end{minipage}
    \begin{minipage}[c]{0.25\textwidth}
        \centering \includegraphics[width=\linewidth,trim={0 0 0 5cm},clip]{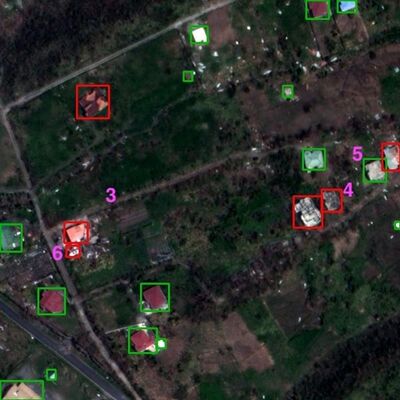}
    \end{minipage}
    \begin{minipage}[c]{0.25\textwidth}
        \centering \includegraphics[width=\linewidth,trim={0 0 0 5cm},clip]{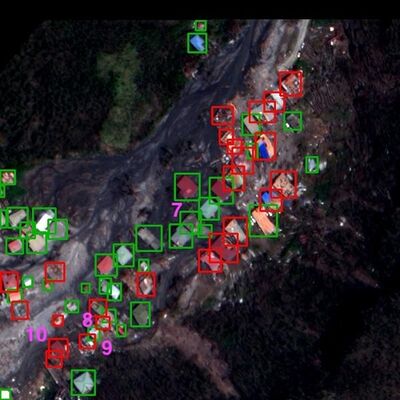}
    \end{minipage}
    \smallskip
    
    \begin{minipage}[c]{0.23\textwidth}
        (e) WBF-EARL
    \end{minipage}
    \begin{minipage}[c]{0.25\textwidth}
        \centering \includegraphics[width=\linewidth,trim={0 0 0 5cm},clip]{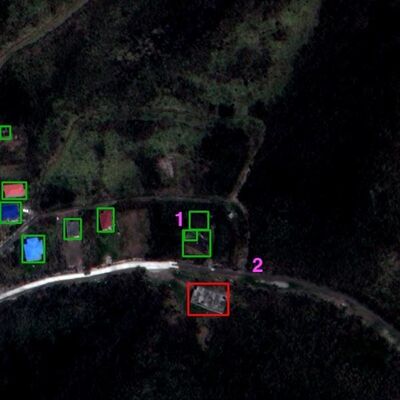}
    \end{minipage}
    \begin{minipage}[c]{0.25\textwidth}
        \centering \includegraphics[width=\linewidth,trim={0 0 0 5cm},clip]{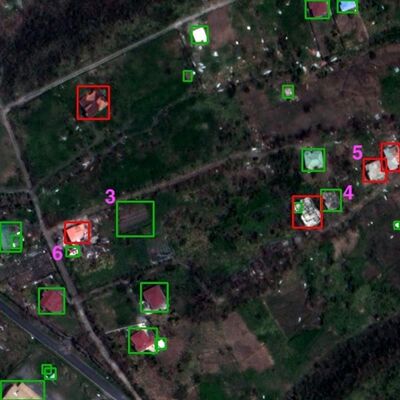}
    \end{minipage}
    \begin{minipage}[c]{0.25\textwidth}
        \centering \includegraphics[width=\linewidth,trim={0 0 0 5cm},clip]{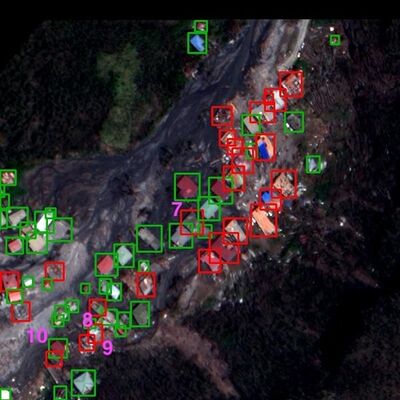}
    \end{minipage}
    \smallskip
    
    \begin{minipage}[c]{0.23\textwidth}
        (f) Crowd R-CNN
    \end{minipage}
    \begin{minipage}[c]{0.25\textwidth}
        \centering \includegraphics[width=\linewidth,trim={0 0 0 5cm},clip]{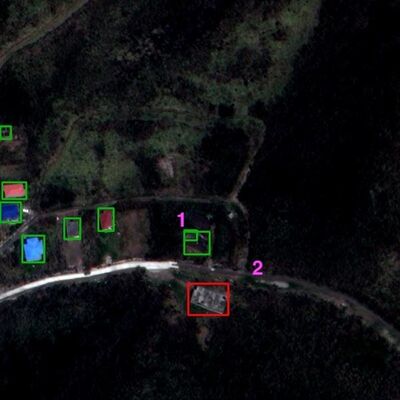}
    \end{minipage}
    \begin{minipage}[c]{0.25\textwidth}
        \centering \includegraphics[width=\linewidth,trim={0 0 0 5cm},clip]{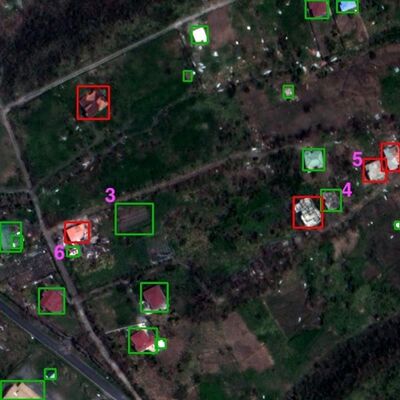}
    \end{minipage}
    \begin{minipage}[c]{0.25\textwidth}
        \centering \includegraphics[width=\linewidth,trim={0 0 0 5cm},clip]{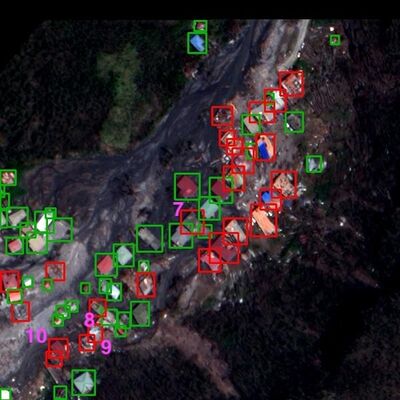}
    \end{minipage}
    \smallskip
    
    \begin{minipage}[c]{0.23\textwidth}
        (g) BDC (ours)
    \end{minipage}
    \begin{minipage}[c]{0.25\textwidth}
        \centering \includegraphics[width=\linewidth,trim={0 0 0 5cm},clip]{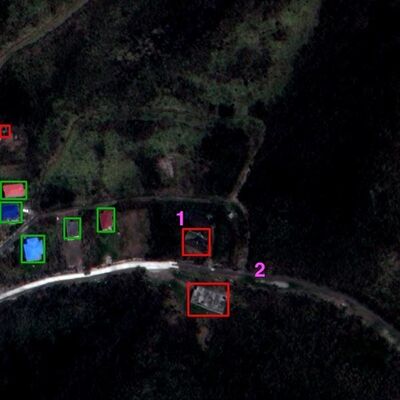}
    \end{minipage}
    \begin{minipage}[c]{0.25\textwidth}
        \centering \includegraphics[width=\linewidth,trim={0 0 0 5cm},clip]{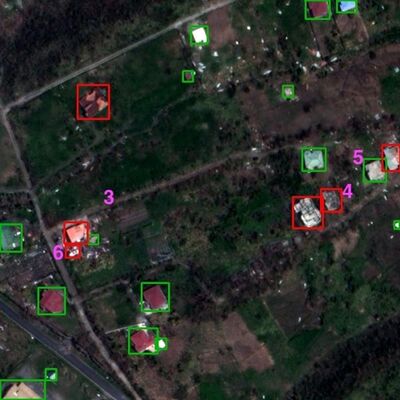}
    \end{minipage}
    \begin{minipage}[c]{0.25\textwidth}
        \centering \includegraphics[width=\linewidth,trim={0 0 0 5cm},clip]{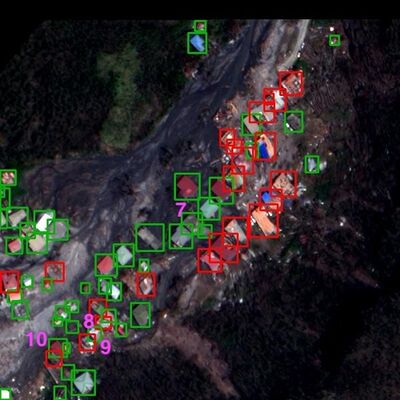}
    \end{minipage}
    \caption{(a)(b) The satellite images before and after the disaster of the same location in the test dataset. (c - g) Predictions of FRCNN object detectors trained via different aggregation methods. The green and red bounding boxes represent the class `undamaged' and `damaged' buildings, respectively. }
    \label{fig:appendix_dis_qual}
\end{figure}

\Cref{fig:appendix_dis_qual} visualises the FRCNN object detectors' predictions trained by different aggregation methods on the test dataset where the original images before and after the disaster are also presented. The images before the disaster are not used to train the object detectors but are used to ascertain whether an object is an undamaged building, a damaged building or not a building. We randomly select three test images representing different degrees of building density for a comprehensive evaluation as shown in \Cref{fig:appendix_dis_qual}. Additionally, we numbered each building of interest and provide brief explanations for each building below.

In the first test example (left column), we observe that Building $1$ is missing a roof (\ie it is damaged) and is only correctly predicted as damaged by FRCNNs trained with NA, MV and BDC. Furthermore, the FRCNN model trained with MV falsely predicts a non-building object (number $2$) as an undamaged building.

For the second test example in the central column, by comparing the images before and after the disaster, we can see that Building $4$ and $6$ are clearly damaged, Building $5$ is undamaged (no significant visual difference between images before and after the disaster), and object number $3$ is not a building. Using this observation as ground truth, the FRCNNs trained via MV and BDC correctly predict all 4 cases while the FRCNNs learnt with other methods did not successfully identify them.

Finally, in the third test case in the right column, it can be observed that Buildings $7$, $9$ and $10$ are likely undamaged, as there are no notable visual differences in the before and after images while Building $8$ is a damaged building. We see that FRCNNs trained with NA, MV, WBF-EARL and Crowd R-CNN  correctly predict 2, 3, 1 and 0 cases, respectively. Moreover, the FRCNNs learnt using WBF-EARL and Crowd R-CNN fail to separate Building $8$ from the building to the right. On the other hand, the FRCNN trained via BDC correctly predict all 4 buildings. They represent some of the empirical evidence which we observe that the FRCNN trained using BDC provides the most accurate predictions.

\clearpage
\section{Synthesisation of Crowdsourced Object Annotations}
\label{appendix:pseudocode_synthesise}
Denote the ground truth annotations for each training image by $\{\Tilde{y}_n\}_{n=1}^{G}$, where $\Tilde{y}_n = \{(s_n, t_n)\}$, $G$ is the number of ground truth instances in the image, $s_n$ and $t_n$ are the $n$-th ground truth bounding box coordinates and class label respectively. Ground truth annotations are only used to synthesise annotations and are not made available to the proposed aggregation method. We use deep learning models to synthesise more realistic crowdsourced annotations with instance-dependent noise. We present the key steps and the pseudocode for the annotation synthesisation procedure as follows.

\paragraph{\bf Class label synthesisation.} Firstly, we use classification networks to obtain a confusion matrix of the ground truth and predicted classes. Images cropped using the ground truth bounding box coordinates and their respective class labels are used to train the classification network. Additionally, we randomly crop regions not covered by ground truth bounding boxes in the images and treat them as the `background' class to simulate false positives and false negatives. The confusion matrix, $\mathcal{C} \in \mathbb{R}^{(J+1)\times (J+1)}$ (the $J+1$ confusion matrix size is due to the added `background' class), obtained by evaluating the classification network on the testing dataset is used as the parameters of Dirichlet distributions so that the $k$-th annotator's confusion matrix, $\mathbf{C}_k$, can be drawn from the same $\mathcal{C}$, \ie $\mathbf{C}_k \sim Dir(\mathcal{C})$. Then, we use $\mathbf{C}_k$ to synthesise class labels from the $k$-th annotator. We do not generate a synthesised annotation if its synthesised class label is `background' to simulate a false negative annotation. 

\paragraph{\bf Bounding box synthesisation.} Next, to synthesise crowdsourced bounding box, $b_m$, we utilise a Region Proposal Network (RPN) \cite{fasterrcnn2015} trained on the ground truth bounding boxes to generate proposals, $\hat{r}$, along with their objectness scores, $\hat{o}$, for a given input image. For each ground truth box, $s_n$, we calculate $\textit{IoU}(s_n, \hat{r}) \odot \hat{o}$, where $\odot$ denotes the element-wise multiplication, and normalise the vector so that it sums to one. Using the normalised vector as the probability vector of a categorical distribution, a proposal is randomly selected as the annotator's bounding box annotation to simulate instance-dependent noise \cite{noisylabelsdnn2022}.

\begin{algorithm}
    \small
    \caption{Synthesising Crowdsourced Object Annotations for the $k$-th Annotator.}
    \label{alg:synthesise}
    \begin{algorithmic}[1]
        \Require{Confusion matrix obtained from the trained classification network, $\mathcal{C} \in \mathbb{R}^{(J+1) \times (J+1)}$, where $J$ is the number of classes; Set of input images, $x=\left\{x^i\in \mathbb{R}^{I_w \times I_h}\right\}_{i=1}^N$, where $I_w$ and $I_h$ are the width and height of the image, respectively, while $N$ is the number of images; Set of ground truth annotations, $\Tilde{y}=\big\{\Tilde{y}^i = \left\{\Tilde{y}^i_n = (s^i_n, t^i_n) \right\}_{n=1}^{G^i}\big\}_{i=1}^N$, where $\Tilde{y}^i_n$ contains the ground truth bounding box coordinates $s^i_n \in \mathbb{R}^4$ and class label $t^i_n \in \{1,...,J\}$ of the $n$-th object in the $i$-th image, while $G^i$ is the number of ground truth objects in the $i$-th image; Trained region proposal networks (RPN), $\textit{RPN}(\cdot):\mathbb{R}^{I_w \times I_h} \to R^L$, where $R=\mathbb{R}^4 \times [0,1]$ is the proposal box coordinates and objectness score while $L$ is the number of proposals; Coverage percentage of the  $k$-th annotator, $\mathfrak{c}_k \in [0, 1]$.}
            \State $\mathbf{C}_k \sim Dir(\mathcal{C})$; \Comment{$\mathcal{C}$ is used as the parameters of Dirichlet distributions to generate $\mathbf{C}_k$.}
            \State $I \gets$ \Call{Sample}{$N, \mathfrak{c}_k$}; \Comment{Sample $\mathfrak{c}_k N$ indices without replacement.}
            \State $y \gets \emptyset$;
            \For{$i$ in $I$}
                \State $y^i \gets \emptyset$;
                \State $\Hat{r}, \Hat{o} \gets$ \textit{RPN}($x^i$); \Comment{Forward pass of RPN to generate proposals.}
                \For{$n=1$ to $G^i$}
                    \State $c^i_n \sim Categorical(\mathbf{C}_{k(t^i_n)})$;
                    \If{$c^i_n = J+1$} \Comment{$J+1$ refers to the augmented `background' class.}
                        \State \textbf{continue} \Comment{Simulate a false negative annotation.}
                    \EndIf
                    \State $p^i_n \gets $ \Call{IoU}{$s^i_n, \Hat{r}$} $\odot\ \Hat{o}$; \Comment{Compute intersection over union (IoU).}
                    \State $p^i_n := \frac{p^i_n}{\lVert p^i_n \rVert_1}$; \Comment{Normalising step so that $p^i_n$ sums to one.}
                    \State $l \sim Categorical(p^i_n)$;
                    \State $b^i_n \gets \Hat{r}_l$; \Comment{Sample a proposal.}
                    \State $y^i := y^i \cup \{(b^i_n, c^i_n, k)\}$;
                \EndFor
                \State $c^i_n \sim Categorical(\mathbf{C}_{k(J+1)})$;
                \If{$c^i_n \neq J+1$} \Comment{Simulate a false positive annotation.}
                    \State $l \sim Categorical(\frac{\Hat{o}}{\lVert \Hat{o} \rVert_1})$;
                    \State $b^i_n \gets \Hat{r}_l$;
                    \State $y^i := y^i \cup \{(b^i_n, c^i_n, k)\}$;
                \EndIf
                \State $y := y \cup y^i$;
            \EndFor
        \Ensure{Synthetic crowdsourced object annotations, $y$.}
    \end{algorithmic}
\end{algorithm}

\paragraph{\bf Realistic crowdsourcing simulation.} See \etal~\cite{crowdquality2013} and Bechtel \etal~\cite{dataquality2017} demonstrated that in real-world scenarios where data annotations are crowdsourced, the annotators often exhibit varying levels of accuracy and consistency. We simulate this by using different deep learning network architectures with distinct levels of performance. Three classification network architectures are used to synthesise crowdsourced class labels, namely AlexNet \cite{alexnet}, VGG16 \cite{vgg16} and ResNet50 \cite{resnet} with an overall test accuracy of 34.65, 55.15, and 73.64 on MS COCO, respectively. On the other hand, the region proposal networks (RPN) with ResNet18, ResNet50 and ResNet101 backbones with respective overall average recall of 45.04, 52.45 and 58.14 on MS COCO, are used to synthesise crowdsourced bounding boxes. 
The deep learning models are trained on MS COCO where the classification networks are trained for 5 epochs with a batch size of 256 and a learning rate of 0.005 while the RPN are trained for 5 epochs with a batch size of 16 and a learning rate of 0.001.
The three pairs of classification networks and RPN backbones simulate annotators of poor, average and expert skill levels described in \Cref{sec:syn_datasets}, respectively.

Furthermore, we simulate false positive annotations for each annotator by randomly synthesising a foreground class label based on $\mathbf{C}_k$ and randomly selecting a bounding box from the proposals. Lastly, to simulate the annotation practice with a large-scale dataset, the synthesisation method has an additional parameter: coverage, which indicates the percentage of images seen by each annotator.

\paragraph{\bf Pseudocode.}\Cref{alg:synthesise} provides the pseudocode to synthesise crowdsourced object annotations for a single annotator, \ie, the $k$-th annotator. In practice, \Cref{alg:synthesise} will be called for all $K$ annotators and their returned results are combined to form the synthetic crowdsourced annotations.

\newpage
\section{Visualisation of Classification Networks and RPN Networks}
\label{appendix:visualise_networks}
To inspect the classification networks and RPN networks discussed in \Cref{appendix:pseudocode_synthesise}, we
visualise in \Cref{fig:conf_matrix,fig:rpn_proposals} the confusion matrices $\mathcal{C}$ of the different classification networks and RPN proposals from the different backbones that are used to synthesise the COCO-MIX dataset.
Small regions of bright spots can be seen around the diagonal of the confusion matrices of the classification networks in \Cref{fig:conf_matrix}. This is because the class indexes of MS COCO are ordered by the class's super-category and objects from classes of the same super-category are usually visually closer to each other, leading to higher correlations among them. For example, class `fork', `knife' and `spoon' with class indices 43, 44 and 45 have the same super-category `kitchen' and are harder to distinguish among them compared to other classes such as `car' or `chair'. 
This shows that a classification network can generate a more realistic confusion matrix $\mathbf{C}$ compared to prior methods which do not take into account the instance-dependent noise, as they set the diagonal of the confusion matrices to an annotator's proficiency level and set the off-diagonal entries either uniformly or randomly \cite{whitehillglad, Raykar2010, crowdlayer2018}. 
Moreover, we observe a clear distinction between the confusion matrices in \Cref{fig:conf_matrix}, simulating annotators of different skills and accuracies.

Similar observations are seen for the RPN proposals. As shown in \Cref{fig:rpn_proposals}, we observe that easily identifiable objects, \eg large television on the right, have more proposals with larger objectness scores compared to the harder, smaller objects. Furthermore, ResNet101 generates more accurate proposals with overall larger objectness scores compared to the ResNet50 and ResNet18 backbones, simulating annotators of different bounding box accuracies.

\begin{figure}[hb]
    \centering
    \begin{subfigure}[b]{0.243\columnwidth}
         \centering
         \includegraphics[width=\linewidth]{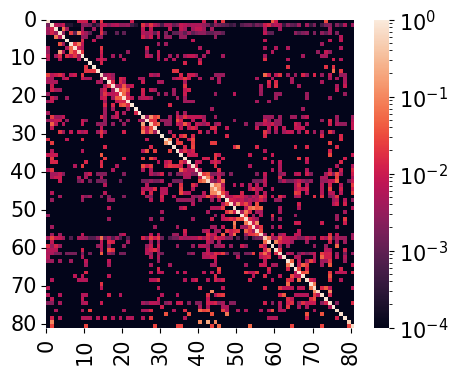}
         \caption{ResNet50}
     \end{subfigure}
    \begin{subfigure}[b]{0.243\columnwidth}
         \centering
         \includegraphics[width=\linewidth]{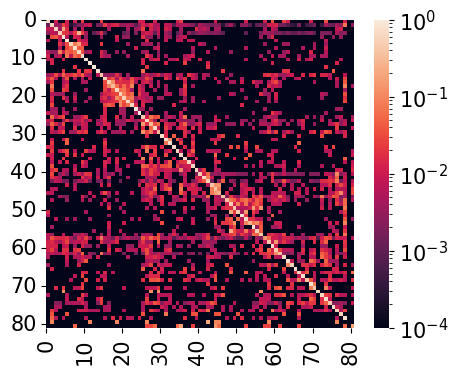}
         \caption{VGG16}
     \end{subfigure}
     \begin{subfigure}[b]{0.243\columnwidth}
         \centering
         \includegraphics[width=\linewidth]{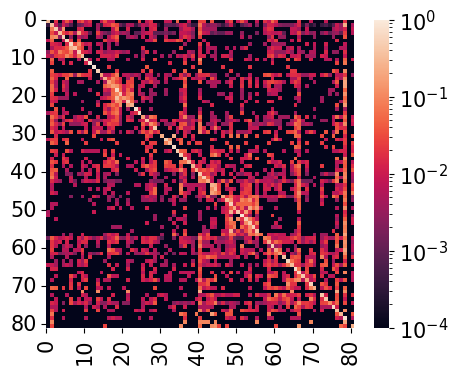}
         \caption{AlexNet}
     \end{subfigure}
    \begin{subfigure}[b]{0.243\columnwidth}
         \centering
         \includegraphics[width=\linewidth]{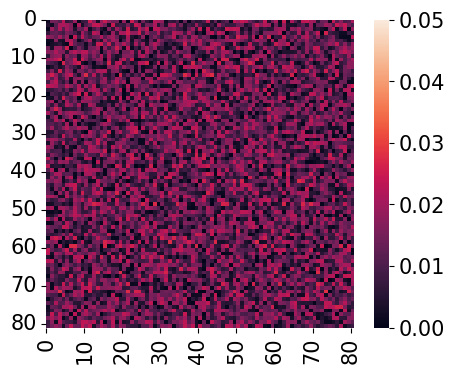}
         \caption{Uniform}
     \end{subfigure}
     \caption{(a) (b) (c) Confusion matrix $\mathcal{C}$ of different network architectures evaluated on the MS COCO test set represented with logarithm-scale heatmaps. (d) Visualisation of the confusion matrix generated by sampling from a uniform distribution (representing the random annotators of COCO-MIX). X-axis is the predicted class while y-axis is the ground truth class. Each row is normalised for better visualisation.}
    \label{fig:conf_matrix}
\end{figure}

\begin{figure}[tb]
    \centering
    \begin{subfigure}[b]{0.3\columnwidth}
         \centering
         \includegraphics[width=\linewidth]{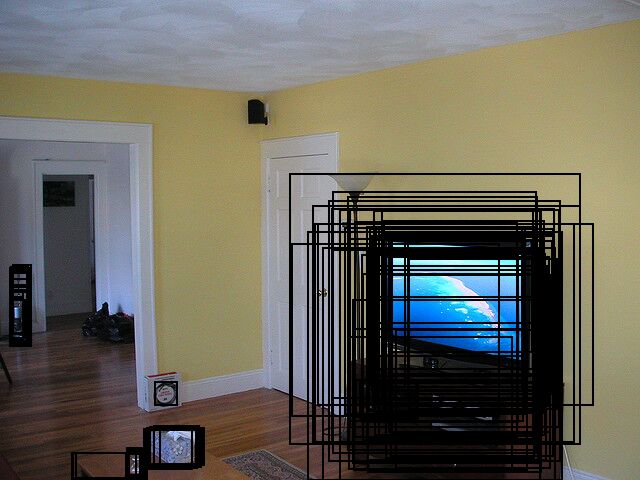}
         \caption{ResNet101}
     \end{subfigure}
     \hfill
     \begin{subfigure}[b]{0.3\columnwidth}
         \centering
         \includegraphics[width=\linewidth]{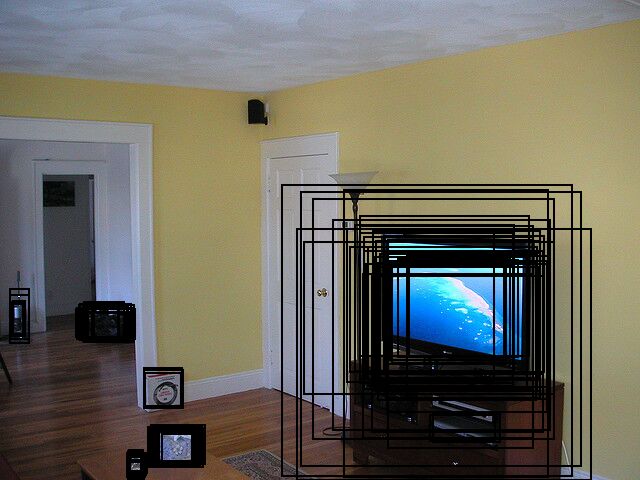}
         \caption{ResNet50}
     \end{subfigure}
     \hfill
     \begin{subfigure}[b]{0.3\columnwidth}
         \centering
         \includegraphics[width=\columnwidth]{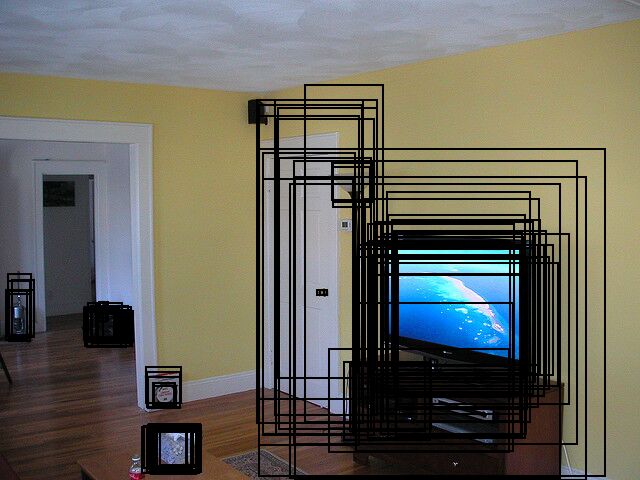}
         \caption{ResNet18}
     \end{subfigure}
     \caption{Top 100 proposals generated by RPN with different backbones, which are then used to synthesise crowdsourced bounding boxes. The brighter box indicates a larger objectness score.}
    \label{fig:rpn_proposals}
\end{figure}

\clearpage
\section{Extensions of BDC to Object Detectors as Annotators} 
\label{appendix:bdc_extension}
Although our work mainly experimented with crowdsourced datasets annotated by real or simulated annotators, the proposed BDC can be extended to scenarios where pre-trained object detector models are directly used as annotators. We synthesise a synthetic dataset utilising COCO-pretrained SSD, SSDLite and FRCNN with MobileNetV3 320 backbone provided by Torchvision. Individually, the 3 models obtain 50.3, 46.7 and 47.8 AP$^{.5:.95}$ on the VOC test dataset. We directly used their object predictions (\ie predicted class labels and bounding boxes) of the VOC training dataset as crowdsourced annotations. 

\Cref{table:weak_obj_metric} summarises the experiment results of YOLOv7, FRCNN and EVA trained with different aggregation methods on said synthetic dataset. We observe that the proposed BDC leads to AP$^{.5:.95}$ of 55.5, 53.0 and 65.1 for YOLOv7, FRCNN and EVA, respectively, on the test dataset which is higher than using MV (52.5, 51.3 and 64.6 for YOLOv7, FRCNN and EVA, respectively). Moreover, BDC consistently achieves the highest AP of the aggregated annotations for the training dataset.

\begin{table}[htbp]
    \caption{Test and train aggregation AP results for YOLOv7, FRCNN and EVA model trained with different annotation aggregation methods on the synthetic crowdsourced VOC dataset created using 3 pre-trained object detectors.}
    \label{table:weak_obj_metric}
    \centering
    \setlength{\tabcolsep}{5pt}
    \resizebox{.9\columnwidth}{!}{
        \begin{tabular}{l|l||ccc|ccc}
            \hline
             \multirow{2}{*}{Model} & \multirow{2}{*}{Method} & \multicolumn{3}{c|}{Test AP} & \multicolumn{3}{c}{Train Aggregation AP} \\
             & & AP$^{.5}$ & AP$^{.75}$ & AP$^{.5:.95}$ & AP$^{.5}$ & AP$^{.75}$ & AP$^{.5:.95}$ \\
             \hline
              \multirow{4}{*}{YOLOv7} & NA & 73.8 & 53.7 & 50.9 & 20.5 & 15.1 & 13.8 \\
              & MV & 78.4 & 58.6 & 52.5 & 54.2 & 38.2 & 34.8 \\
              & WBF-EARL \cite{Le2023} & 75.8 & 59.7 & 52.7 & 54.6 & 32.2 & 32.1 \\
              & \textbf{BDC (ours)} & \textbf{81.9} & \textbf{63.2} & \textbf{55.5} & \textbf{72.5} & \textbf{51.9} & \textbf{47.3} \\
              \hline
              \multirow{5}{*}{FRCNN} & NA & 71.7 & 47.6 & 44.6 & 20.5 & 15.1 & 13.8 \\
              & MV & 77.0 & 58.0 & 51.3 & 54.2 & 38.2 & 34.8 \\
              & Crowd R-CNN \cite{crowdrcnn2020} & 75.4 & 54.4 & 50.7 & 52.4 & 33.1 & 32.6 \\
              & WBF-EARL \cite{Le2023}& 79.1 & 55.1 & 50.2 & 54.6 & 32.2 & 32.1 \\
              & \textbf{BDC (ours)} & \textbf{81.0} & \textbf{59.5} & \textbf{53.0} & \textbf{71.3} & \textbf{48.6} & \textbf{45.5} \\
              \hline
              \multirow{4}{*}{EVA} & NA & 86.3 & 74.5 & 63.1 & 20.5 & 15.1 & 13.8 \\
              & MV & 87.7 & \textbf{75.6} & 64.7 & 54.2 & 38.2 & 34.8 \\
              & WBF-EARL \cite{Le2023}& 87.5 & 75.2 & 64.4 & 54.6 & 32.2 & 32.1 \\
              & \textbf{BDC (ours)} & \textbf{87.9} & 75.4 & \textbf{65.1} & \textbf{71.7} & \textbf{48.4} & \textbf{45.6} \\
              \hline
        \end{tabular}
    }
\end{table}

\clearpage
\section{Additional Qualitative Results}
\begin{figure}[hb]
    \captionsetup[subfigure]{format=hang}

    \begin{subfigure}{0.16\textwidth}
        \centering \includegraphics[width=\linewidth]{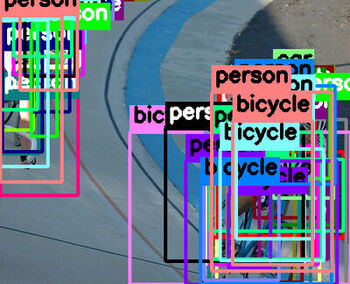}
    \end{subfigure}
    \begin{subfigure}{0.16\textwidth}
        \centering \includegraphics[width=\linewidth]{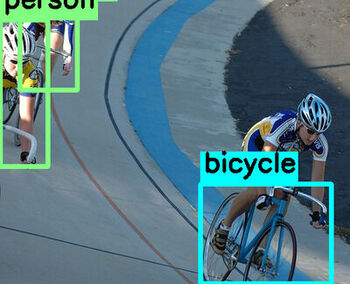}
    \end{subfigure}
    \begin{subfigure}{0.16\textwidth}
        \centering \includegraphics[width=\linewidth]{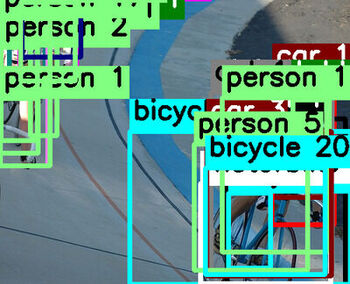}
    \end{subfigure}
    \begin{subfigure}{0.16\textwidth}
        \centering \includegraphics[width=\linewidth]{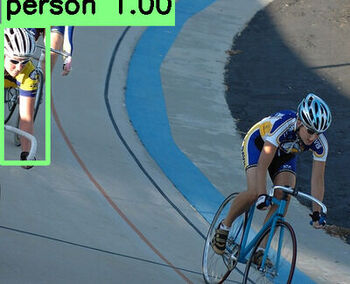}
    \end{subfigure}
    \begin{subfigure}{0.16\textwidth}
        \centering \includegraphics[width=\linewidth]{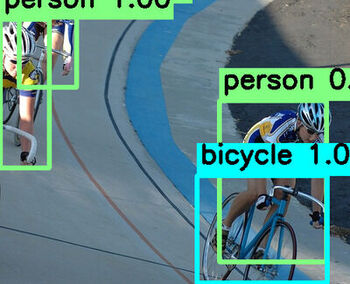}
    \end{subfigure}
    \begin{subfigure}{0.16\textwidth}
        \centering \includegraphics[width=\linewidth]{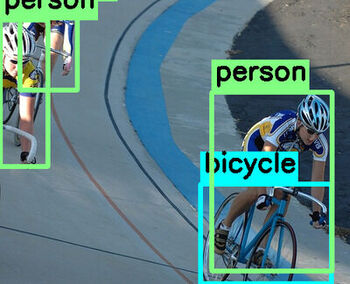}
    \end{subfigure}
    
    \smallskip
    \begin{subfigure}{0.16\textwidth}
        \centering \includegraphics[width=\linewidth,trim={3cm 0cm 0cm 3cm},clip]{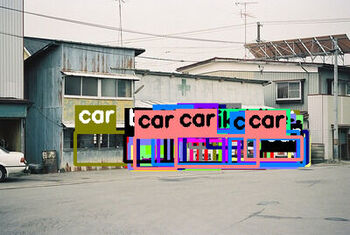}
    \end{subfigure}
    \begin{subfigure}{0.16\textwidth}
        \centering \includegraphics[width=\linewidth,trim={3cm 0cm 0cm 3cm},clip]{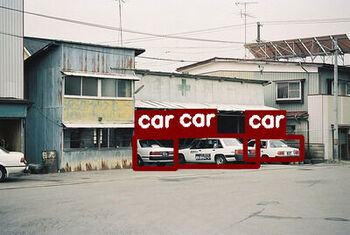}
    \end{subfigure}
    \begin{subfigure}{0.16\textwidth}
        \centering \includegraphics[width=\linewidth,trim={3cm 0cm 0cm 3cm},clip]{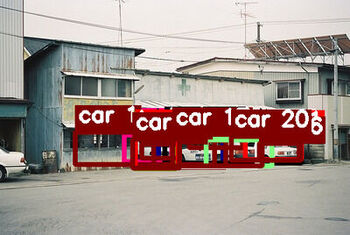}
    \end{subfigure}
    \begin{subfigure}{0.16\textwidth}
        \centering \includegraphics[width=\linewidth,trim={3cm 0cm 0cm 3cm},clip]{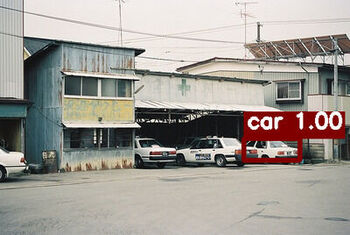}
    \end{subfigure}
    \begin{subfigure}{0.16\textwidth}
        \centering \includegraphics[width=\linewidth,trim={3cm 0cm 0cm 3cm},clip]{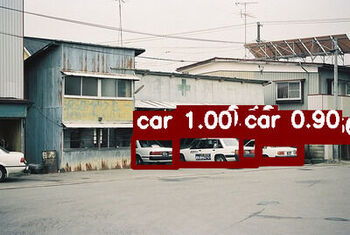}
    \end{subfigure}
    \begin{subfigure}{0.16\textwidth}
        \centering \includegraphics[width=\linewidth,trim={3cm 0cm 0cm 3cm},clip]{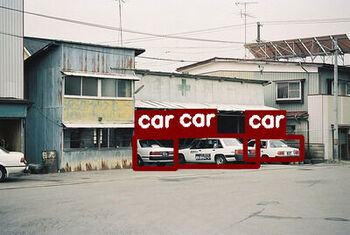}
    \end{subfigure}

    \smallskip
    \begin{subfigure}{0.16\textwidth}
        \centering \includegraphics[width=\linewidth,trim={3cm 0cm 0cm 3cm},clip]{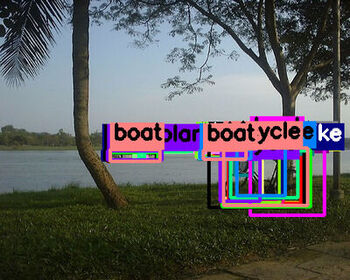}
    \end{subfigure}
    \begin{subfigure}{0.16\textwidth}
        \centering \includegraphics[width=\linewidth,trim={3cm 0cm 0cm 3cm},clip]{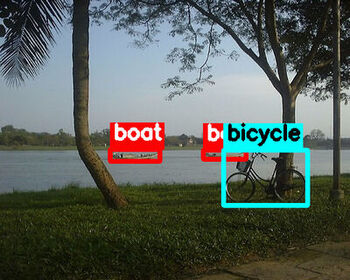}
    \end{subfigure}
    \begin{subfigure}{0.16\textwidth}
        \centering \includegraphics[width=\linewidth,trim={3cm 0cm 0cm 3cm},clip]{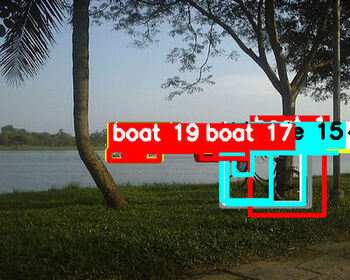}
    \end{subfigure}
    \begin{subfigure}{0.16\textwidth}
        \centering \includegraphics[width=\linewidth,trim={3cm 0cm 0cm 3cm},clip]{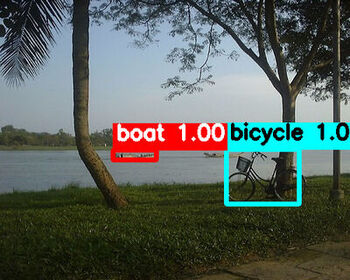}
    \end{subfigure}
    \begin{subfigure}{0.16\textwidth}
        \centering \includegraphics[width=\linewidth,trim={3cm 0cm 0cm 3cm},clip]{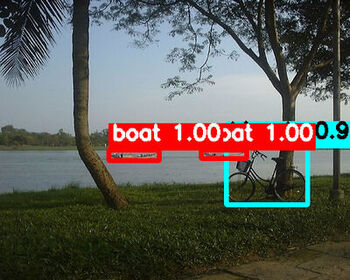}
    \end{subfigure}
    \begin{subfigure}{0.16\textwidth}
        \centering \includegraphics[width=\linewidth,trim={3cm 0cm 0cm 3cm},clip]{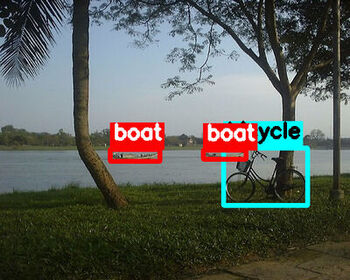}
    \end{subfigure}

    \smallskip
    \begin{subfigure}{0.16\textwidth}
        \centering \includegraphics[width=\linewidth]{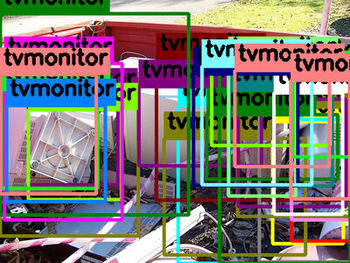}
    \end{subfigure}
    \begin{subfigure}{0.16\textwidth}
        \centering \includegraphics[width=\linewidth]{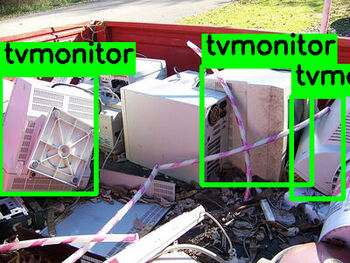}
    \end{subfigure}
    \begin{subfigure}{0.16\textwidth}
        \centering \includegraphics[width=\linewidth]{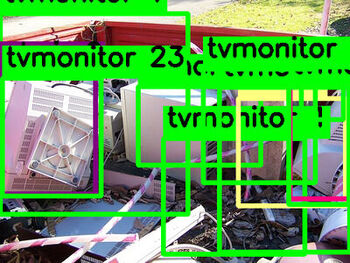}
    \end{subfigure}
    \begin{subfigure}{0.16\textwidth}
        \centering \includegraphics[width=\linewidth]{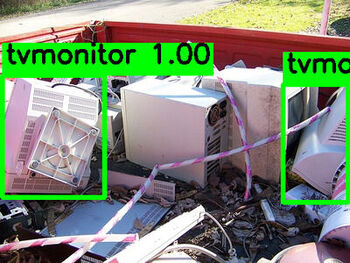}
    \end{subfigure}
    \begin{subfigure}{0.16\textwidth}
        \centering \includegraphics[width=\linewidth]{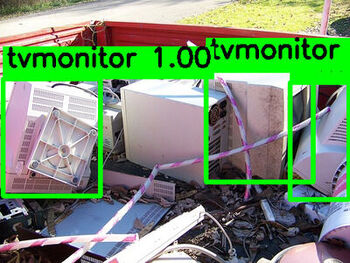}
    \end{subfigure}
    \begin{subfigure}{0.16\textwidth}
        \centering \includegraphics[width=\linewidth]{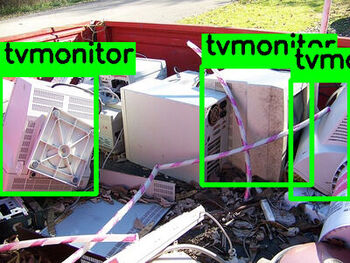}
    \end{subfigure}

    \smallskip
    \begin{subfigure}{0.16\textwidth}
        \centering \includegraphics[width=\linewidth]{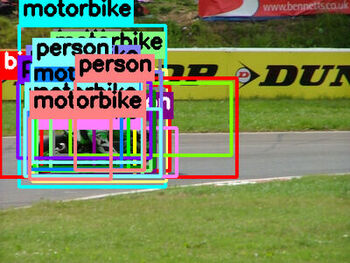}
    \end{subfigure}
    \begin{subfigure}{0.16\textwidth}
        \centering \includegraphics[width=\linewidth]{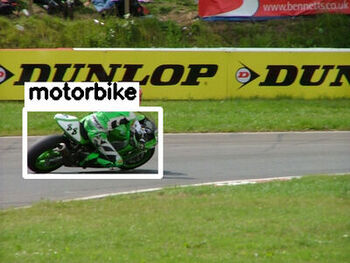}
    \end{subfigure}
    \begin{subfigure}{0.16\textwidth}
        \centering \includegraphics[width=\linewidth]{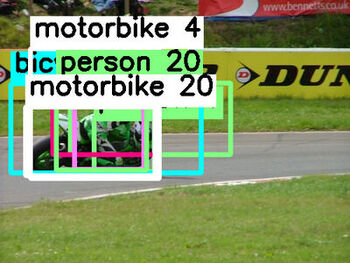}
    \end{subfigure}
    \begin{subfigure}{0.16\textwidth}
        \centering \includegraphics[width=\linewidth]{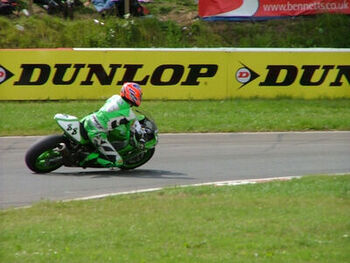}
    \end{subfigure}
    \begin{subfigure}{0.16\textwidth}
        \centering \includegraphics[width=\linewidth]{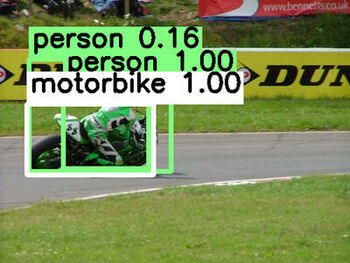}
    \end{subfigure}
    \begin{subfigure}{0.16\textwidth}
        \centering \includegraphics[width=\linewidth]{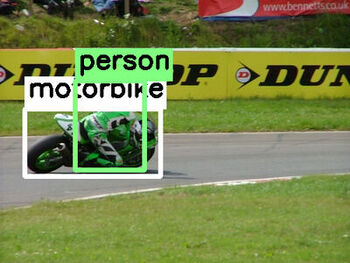}
    \end{subfigure}

    \smallskip

    \subcaptionbox{NA}{\includegraphics[width=0.16\textwidth]{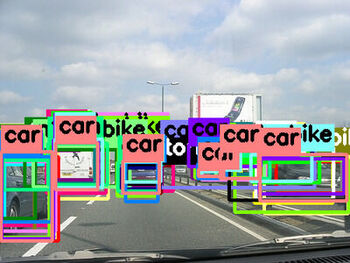}}
    \subcaptionbox{MV}{\includegraphics[width=0.16\textwidth]{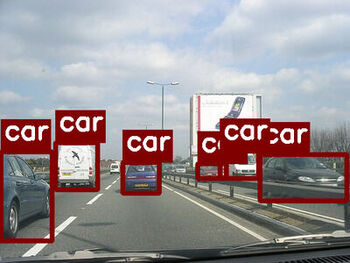}}
    \subcaptionbox{WBF-EARL}{\includegraphics[width=0.16\textwidth]{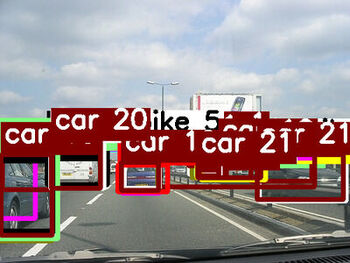}}
    \subcaptionbox{Crowd R-CNN}{\includegraphics[width=0.16\textwidth]{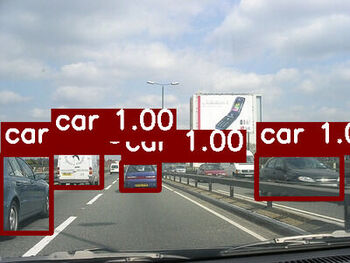}}
    \subcaptionbox{BDC (ours)}{\includegraphics[width=0.16\textwidth]{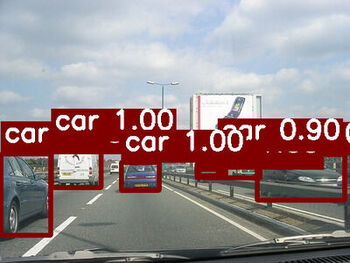}}
    \subcaptionbox{Ground truth}{\includegraphics[width=0.16\textwidth]{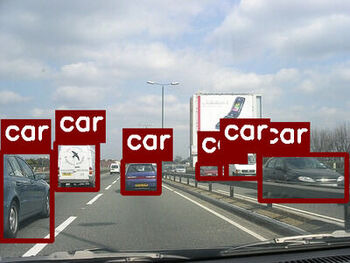}}
    
    \caption{Additional comparisons of the aggregated labels from different methods on the VOC-MIX dataset. For WBF-EARL \cite{Le2023}, the number beside the class label is the annotators' level of agreement while for Crowd R-CNN \cite{crowdrcnn2020} and BDC, the number indicates the class probability. For NA, the colours represent the different annotators.}
    \label{fig:appendix_add_qual}
\end{figure}

\Cref{fig:appendix_add_qual} shows additional results of the aggregated annotations from each method on the VOC-MIX dataset. The proposed BDC outperforms other methods across various types of objects, \eg small objects, objects with overlapping bounding boxes, \etc.
\label{appendix:add_qualitative}

\newpage
\section{Analysis of Posterior Distributions} 
\label{appendix:posterior_analysis}
\begin{figure}[hb]
    \begin{subfigure}[b]{0.245\columnwidth}
         \centering
         \includegraphics[width=1\linewidth]{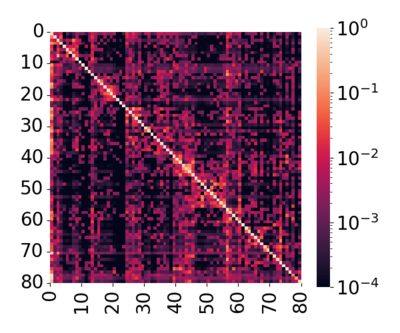}
         \caption{C1 (0.692)}
     \end{subfigure}
     \begin{subfigure}[b]{0.245\columnwidth}
         \centering
         \includegraphics[width=1\linewidth]{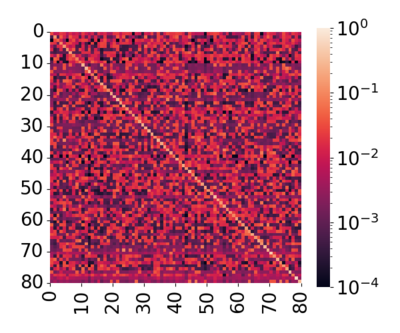}
         \caption{C2 (0.174)}
     \end{subfigure}
      \begin{subfigure}[b]{0.245\columnwidth}
         \centering
         \includegraphics[width=1\linewidth]{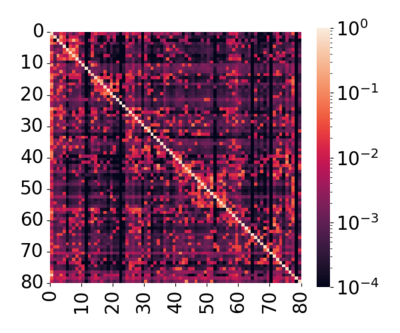}
         \caption{C3 (0.599)}
     \end{subfigure}
      \begin{subfigure}[b]{0.245\columnwidth}
         \centering
         \includegraphics[width=1\linewidth]{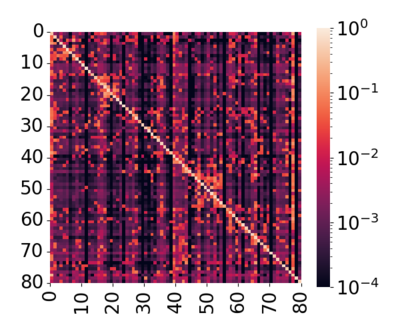}
         \caption{C4 (0.387)}
     \end{subfigure}
    \caption{The mean of the Dirichlet posteriors $\pi^k$ of the CLA of the COCO-MIX YOLOv7 experiment visualised as a logarithm-scale heatmap for each cluster found using K-means clustering \cite{kmeans} on the aggregators' parameters (\ie $\bm{\mu}$, $\bm{\upsilon}$, $\bm{\beta}$ and $\bm{\alpha}$). X-axis is the annotated class while y-axis is the ground truth class. The average diagonal value of each cluster's $\pi^k$ is indicated in brackets in each caption.}
    \label{fig:agg_cm}
\end{figure}

\begin{figure}[tb]
     \centering
     \includegraphics[width=0.6\columnwidth]{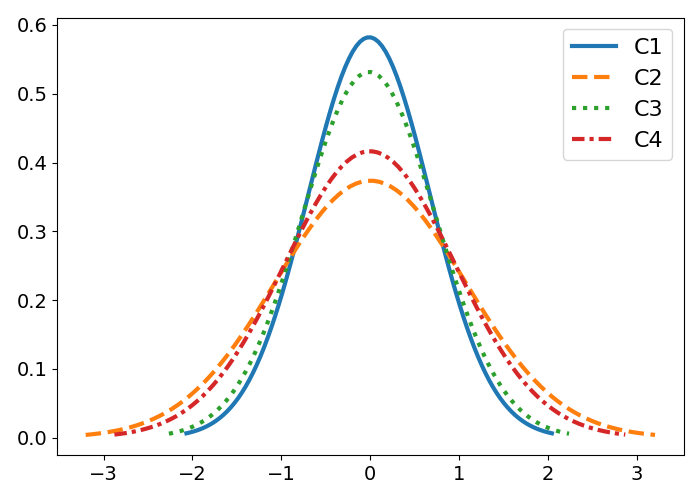}
     \caption{The Gaussian distributions for the x-axis translation error of the BBA of the COCO-MIX YOLOv7 experiment for each cluster found using K-means clustering \cite{kmeans} on the aggregators' parameters (\ie $\bm{\mu}$, $\bm{\upsilon}$, $\bm{\beta}$ and $\bm{\alpha}$).}
    \label{fig:agg_gauss}
\end{figure}

The proposed BDC enables the grouping of the annotators into different clusters of similar skill levels by analysing the learnt distributions of bounding box errors and confusion matrices for each annotator. We perform a K-means clustering \cite{kmeans} on the parameters of these distributions, \ie $\bm{\mu}$, $\bm{\upsilon}$, $\bm{\beta}$ and $\bm{\alpha}$, of the COCO-MIX YOLOv7 experiment by setting the number of clusters to four. \Cref{fig:agg_cm,fig:agg_gauss} visualise the mean of each cluster's Dirichlet posteriors $\pi^k$ of the CLA and the Gaussian distribution for the x-axis translation error of BBA, respectively. We observe that the clusters of annotators correspond to the four groups of annotators in the COCO-MIX synthetic setting discussed in \Cref{sec:syn_datasets}, \ie C1, C3, C4 and C2 correspond to expert, average, poor and random annotators, respectively.

For example, the average diagonal values of $\pi^k$ for clusters C1, C3 and C4 are similar to the overall test accuracy of the three classification networks, \ie ResNet50, VGG16 and AlexNet, used to synthesise the crowdsourced class labels, while the heatmap of C2 is close to a uniform distribution; The variances for the x-axis translation error of each cluster are consistent with those of the four groups of annotators in the COCO-MIX synthetic setting (\ie variance of $\text{C1} < \text{C3} < \text{C4} < \text{C2}$). This exemplifies the ability of the BDC framework to correctly learn the annotators' skill levels. 

%% file: main.bbl
\begin{thebibliography}{10}
\providecommand{\url}[1]{\texttt{#1}}
\providecommand{\urlprefix}{URL }
\providecommand{\doi}[1]{https://doi.org/#1}

\bibitem{gammafunc}
Abramowitz, M., Stegun, I.A.: Handbook of Mathematical Functions with Formulas,
  Graphs, and Mathematical Tables, chap.~6, pp. 255--258. Dover, Mineola, NY,
  USA, 10 edn. (1972)

\bibitem{digamma}
Abramowitz, M., Stegun, I.A.: Handbook of Mathematical Functions with Formulas,
  Graphs, and Mathematical Tables, chap.~6, pp. 258--259. Dover, Mineola, NY,
  USA, 10 edn. (1972)

\bibitem{aggnet2016}
Albarqouni, S., Baur, C., Achilles, F., Belagiannis, V., Demirci, S., Navab,
  N.: {AggNet}: Deep learning from crowds for mitosis detection in breast
  cancer histology images. IEEE Trans. Med. Imag.  \textbf{35}(5),  1313--1321
  (May 2016)

\bibitem{phiseg2019}
Baumgartner, C.F., Tezcan, K.C., Chaitanya, K., H{\"o}tker, A.M., Muehlematter,
  U.J., Schawkat, K., Becker, A.S., Donati, O., Konukoglu, E.: {PHiSeg}:
  Capturing uncertainty in medical image segmentation. In: Med. Imag. Comput.
  Comput. Assist. Interv. pp. 119--127. Shenzhen, China (2019)

\bibitem{dataquality2017}
Bechtel, B., Demuzere, M., Sismanidis, P., Fenner, D., Brousse, O., Beck, C.,
  Van~Coillie, F., Conrad, O., Keramitsoglou, I., Middel, A., Mills, G.,
  Niyogi, D., Otto, M., See, L., Verdonck, M.L.: Quality of crowdsourced data
  on urban morphology—the human influence experiment ({HUMINEX}). Urban Sci.
  \textbf{1}(2) (May 2017)

\bibitem{gaussgamma}
Bernard, J.M., Smith, A.F.M.: Bayesian Theory, chap.~3, pp. 136--138. Wiley,
  West Sussex, UK (1993)

\bibitem{activelabel}
Bernhardt, M., Castro, D.C., Tanno, R., Schwaighofer, A., Tezcan, K.C.,
  Monteiro, M., Bannur, S., Lungren, M.P., Nori, A., Glocker, B.,
  Alvarez-Valle, J., Oktay, O.: Active label cleaning for improved dataset
  quality under resource constraints. Nature Commun.  \textbf{13}(1), ~1161
  (Mar 2022)

\bibitem{canbudd2021}
Budd, S., Day, T., Simpson, J., Lloyd, K., Matthew, J., Skelton, E., Razavi,
  R., Kainz, B.: Can non-specialists provide high quality gold standard labels
  in challenging modalities? In: Med. Imag. Comput. Comput. Assist. Interv. pp.
  251--262. Strasbourg, France (2021)

\bibitem{detr2020}
Carion, N., Massa, F., Synnaeve, G., Usunier, N., Kirillov, A., Zagoruyko, S.:
  End-to-end object detection with transformers. In: Eur. Conf. Comput. Vis.
  pp. 213--229 (2020)

\bibitem{chenlabelsep}
Chen, Z., Wang, H., Sun, H., Chen, P., Han, T., Liu, X., Yang, J.: Structured
  probabilistic end-to-end learning from crowds. In: Int. Joint Conf. Artif.
  Intell. pp. 1512--1518 (2020)

\bibitem{davidskene1979}
Dawid, A.P., Skene, A.M.: Maximum likelihood estimation of observer error-rates
  using the {EM} algorithm. J. Roy. Statist. Soc. Ser. C (Appl. Statist.)
  \textbf{28}(1),  20--28 (1979)

\bibitem{vindrkaggle}
DungNB, Nguyen, H.Q., Elliott, J., KeepLearning, Nhan, N.T., Culliton, P.:
  {VinBigData} chest x-ray abnormalities detection (2020),
  \url{https://kaggle.com/competitions/vinbigdata-chest-xray-abnormalities-detection}

\bibitem{pascal-voc-2007}
Everingham, M., Luc Van~Gool, C.K.I.W., Winn, J., Zisserman, A.: The {PASCAL}
  {V}isual {O}bject {C}lasses {C}hallenge 2007 {(VOC2007)}. Int. J. Comput.
  Vis.  \textbf{88},  303--338 (Jun 2010)

\bibitem{eva2023}
Fang, Y., Wang, W., Xie, B., Sun, Q., Wu, L., Wang, X., Huang, T., Wang, X.,
  Cao, Y.: {EVA}: Exploring the limits of masked visual representation learning
  at scale. In: IEEE Conf. Comput. Vis. Pattern Recog. pp. 19358--19369.
  Vancouver, Canada (2023)

\bibitem{gohcrowdlab}
Goh, H.W., Tkachenko, U., Mueller, J.: {CROWDLAB}: Supervised learning to infer
  consensus labels and quality scores for data with multiple annotators (2022),
  arXiv:2210.06812

\bibitem{guanlabelsep}
Guan, M.Y., Gulshan, V., Dai, A.M., Hinton, G.E.: Who said what: Modeling
  individual labelers improves classification. In: AAAI Conf. Artif. Intell.
  pp. 3109--3118. San Francisco, CA, USA (2017)

\bibitem{resnet}
He, K., Zhang, X., Ren, S., Sun, J.: Deep residual learning for image
  recognition. In: IEEE Conf. Comput. Vis. Pattern Recog. pp. 770--778. Las
  Vegas, NV, USA (2016)

\bibitem{crowdrcnn2020}
Hu, Y., Meina, S.: Crowd {R-CNN}: An object detection model utilizing
  crowdsourced labels. In: Int. Conf. Vis. Image Sig. Process. pp.~1--7.
  Bangkok, Thailand (2020)

\bibitem{bccnet2018}
Isupova, O., Li, Y., Kuzin, D., Roberts, S.J., Willis, K.J., Reece, S.:
  {BCCNet}: Bayesian classifier combination neural network. In: Neural Inf.
  Process. Syst. Workshop Mach. Learn. Develop. Montr\'{e}al, Canada (2018)

\bibitem{jensenlabelsep}
Jensen, M.H., J{\o}rgensen, D.R., Jalaboi, R., Hansen, M.E., Olsen, M.A.:
  Improving uncertainty estimation in convolutional neural networks using
  inter-rater agreement. In: Med. Imag. Comput. Comput. Assist. Interv. pp.
  540--548. Shenzhen, China (2019)

\bibitem{multiratersegWei2021}
Ji, W., Yu, S., Wu, J., Ma, K., Bian, C., Bi, Q., Li, J., Liu, H., Cheng, L.,
  Zheng, Y.: Learning calibrated medical image segmentation via multi-rater
  agreement modeling. In: IEEE Conf. Comput. Vis. Pattern Recog. pp.
  12336--12346 (2021)

\bibitem{jungolabelsep}
Jungo, A., Meier, R., Ermis, E., Blatti-Moreno, M., Herrmann, E., Wiest, R.,
  Reyes, M.: On the effect of inter-observer variability for a reliable
  estimation of uncertainty of medical image segmentation. In: Med. Imag.
  Comput. Comput. Assist. Interv. pp. 682--690. Granada, Spain (2018)

\bibitem{BCC2012}
Kim, H., Ghahramani, Z.: Bayesian classifier combination. In: Int. Conf. Artif.
  Intell. Statist. pp. 619--627. La Palma, Canary Islands (2012)

\bibitem{probunet2018}
Kohl, S.A.A., Romera-Paredes, B., Meyer, C., Fauw, J.D., Ledsam, J.R.,
  Maier-Hein, K.H., Eslami, S.M.A., Rezende, D.J., Ronneberger, O.: A
  probabilistic u-net for segmentation of ambiguous images. In: Adv. Neural
  Inform. Process. Syst. pp. 6965--6975. Montr\'{e}al, Canada (2018)

\bibitem{alexnet}
Krizhevsky, A., Sutskever, I., Hinton, G.E.: {ImageNet} classification with
  deep convolutional neural networks. In: Adv. Neural Inform. Process. Syst.
  pp. 1097--1105. Lake Tahoe, NV, USA (2012)

\bibitem{mammointerobs2006}
Lazarus, E., Mainiero, M.B., Schepps, B., Koelliker, S.L., Livingston, L.S.:
  {BI-RADS} lexicon for {US} and mammography: interobserver variability and
  positive predictive value. Radiology  \textbf{239}(2),  385--391 (Mar 2006)

\bibitem{Le2023}
Le, K., Tran, T., Pham, H., Nguyen~Trung, H., Le, T., Nguyen, H.Q.: Learning
  from multiple expert annotators for enhancing anomaly detection in medical
  image analysis. IEEE Access  \textbf{11},  14105--14114 (2023)

\bibitem{mnist2010}
LeCun, Y., Cortes, C., Burges, C.: {MNIST} handwritten digit database (2010),
  \url{http://yann.lecun.com/exdb/mnist}

\bibitem{cocodataset}
Lin, T.Y., Maire, M., Belongie, S., Hays, J., Perona, P., Ramanan, D., Dollar,
  P., Zitnick, L.: Microsoft {COCO}: Common objects in context. In: Eur. Conf.
  Comput. Vis. pp. 740--755. Zurich, Switzerland (2014)

\bibitem{kmeans}
Lloyd, S.P.: Least squares quantization in {PCM}. IEEE Trans. Inf. Theory
  \textbf{28}(2),  129--137 (Mar 1982)

\bibitem{cosdecay}
Loshchilov, I., Hutter, F.: {SGDR}: Stochastic gradient descent with warm
  restarts. In: Int. Conf. Learn. Represent. Toulon, France (2017)

\bibitem{adamw}
Loshchilov, I., Hutter, F.: Decoupled weight decay regularization. In: Int.
  Conf. Learn. Represent. New Orleans, LA, USA (2019)

\bibitem{crowdsourcingLena2014}
Maier-Hein, L., Mersmann, S., Kondermann, D., Stock, C., Kenngott, H.G.,
  Sanchez, A., Wagner, M., Preukschas, A., Wekerle, A.L., Helfert, S.,
  Bodenstedt, S., Speidel, S.: Crowdsourcing for reference correspondence
  generation in endoscopic images. In: Med. Imag. Comput. Comput. Assist.
  Interv. pp. 349--356. Boston, MA, USA (2014)

\bibitem{softlabels}
M\"{u}ller, R., Kornblith, S., Hinton, G.: When does label smoothing help? In:
  Adv. Neural Inform. Process. Syst. pp. 4694--4703. Vancouver, Canada (2019)

\bibitem{normalconj}
Murphy, K.: Conjugate bayesian analysis of the gaussian distribution (2007),
  \url{https://www.cs.ubc.ca/~murphyk/Papers/bayesGauss.pdf}

\bibitem{vindr2022}
Nguyen, H.Q., Lam, K., Le, L.T., Pham, H.H., Tran, D.Q., Nguyen, D.B., Le,
  D.D., Pham, C.M., Tong, H.T.T., Dinh, D.H., Do, C.D., Doan, L.T., Nguyen,
  C.N., Nguyen, B.T., Nguyen, Q.V., Hoang, A.D., Phan, H.N., Nguyen, A.T., Ho,
  P.H., Ngo, D.T., Nguyen, N.T., Nguyen, N.T., Dao, M., Vu, V.: {VinDr-CXR}: An
  open dataset of chest x-rays with radiologist's annotations. Sci. Data
  \textbf{9}(1), ~429 (Jul 2022)

\bibitem{nowruzidataforobjdet}
Nowruzi, F.E., Kapoor, P., Kolhatkar, D., Hassanat, F., Laganiere, R., Rebut,
  J.: How much real data do we actually need: Analyzing object detection
  performance using synthetic and real data. In: Int. Conf. Mach. Learn.
  Workshop AI Auton. Driving. Long Beach, CA, USA (2019)

\bibitem{pytorch}
Paszke, A., Gross, S., Massa, F., Lerer, A., Bradbury, J., Chanan, G., Killeen,
  T., Lin, Z., Gimelshein, N., Antiga, L., Desmaison, A., Kopf, A., Yang, E.,
  DeVito, Z., Raison, M., Tejani, A., Chilamkurthy, S., Steiner, B., Fang, L.,
  Bai, J., Chintala, S.: Pytorch: An imperative style, high-performance deep
  learning library. In: Adv. Neural Inform. Process. Syst. pp. 8024--8035.
  Vancouver, Canada (2019)

\bibitem{Raykar2010}
Raykar, V.C., Yu, S., Zhao, L.H., Valadez, G.H., Florin, C., Bogoni, L., Moy,
  L., Raykar, C., Valadez, G.H.: Learning from crowds. J. Mach. Learn. Res.
  \textbf{11}(43),  1297--1322 (Aug 2010)

\bibitem{yolo}
Redmon, J., Farhadi, A.: {YOLOv3}: An incremental improvement (2018),
  arXiv:1804.02767

\bibitem{fasterrcnn2015}
Ren, S., He, K., Girshick, R., Sun, J.: Faster {R-CNN}: Towards real-time
  object detection with region proposal networks. In: Adv. Neural Inform.
  Process. Syst. pp. 91--99. Montr\'{e}al, Canada (2015)

\bibitem{giou}
Rezatofighi, H., Tsoi, N., Gwak, J., Sadeghian, A., Reid, I., Savarese, S.:
  Generalized intersection over union: A metric and a loss for bounding box
  regression. In: IEEE Conf. Comput. Vis. Pattern Recog. pp. 658--666. Long
  Beach, CA, USA (2019)

\bibitem{crowdlayer2018}
Rodrigues, F., Pereira, F.C.: Deep learning from crowds. In: AAAI Conf. Artif.
  Intell. pp. 1611--1618. New Orleans, LA, USA (2018)

\bibitem{imagenet}
Russakovsky, O., Deng, J., Su, H., Krause, J., Satheesh, S., Ma, S., Huang, Z.,
  Karpathy, A., Khosla, A., Bernstein, M., Berg, A.C., Fei-Fei, L.: {ImageNet}
  large scale visual recognition challenge. Int. J. Comput. Vis.
  \textbf{115}(3),  211--252 (Dec 2015)

\bibitem{crowdquality2013}
See, L., Comber, A., Salk, C., Fritz, S., van~der Velde, M., Perger, C.,
  Schill, C., McCallum, I., Kraxner, F., Obersteiner, M.: Comparing the quality
  of crowdsourced data contributed by expert and non-experts. PLoS One
  \textbf{8}(7),  e69958 (Jul 2013)

\bibitem{majorvote2008}
Sheng, V.S., Provost, F., Ipeirotis, P.G.: Get another label? improving data
  quality and data mining using multiple, noisy labelers. In: ACM SIGKDD Int.
  Conf. Knowl. Discov. Data Min. pp. 614--622. Las Vegas, NV, USA (2008)

\bibitem{vgg16}
Simonyan, K., Zisserman, A.: Very deep convolutional networks for large-scale
  image recognition. In: Int. Conf. Learn. Represent. San Diego, CA, USA (2015)

\bibitem{WBF2021}
Solovyev, R., Wang, W., Gabruseva, T.: Weighted boxes fusion: Ensembling boxes
  from different object detection models. Image Vis. Comput.  \textbf{107},
  104--117 (Mar 2021)

\bibitem{noisylabelsdnn2022}
Song, H., Kim, M., Park, D., Shin, Y., Lee, J.G.: Learning from noisy labels
  with deep neural networks: A survey. IEEE Trans. Neural Netw. Learn. Syst.
  \textbf{34}(11),  8135--8153 (Mar 2022)

\bibitem{agreetodisagree2019}
Sudre, C.H., Anson, B.G., Ingala, S., Lane, C.D., Jimenez, D., Haider, L.,
  Varsavsky, T., Tanno, R., Smith, L., Ourselin, S., J{\"a}ger, R.H., Cardoso,
  M.J.: Let's agree to disagree: Learning highly debatable multirater
  labelling. In: Med. Imag. Comput. Comput. Assist. Interv. pp. 665--673.
  Shenzhen, China (2019)

\bibitem{commbcc2014}
Venanzi, M., Guiver, J., Kazai, G., Kohli, P., Shokouhi, M.: Community-based
  bayesian aggregation models for crowdsourcing. In: Int. Conf. World Wide Web.
  pp. 155--164. Seoul, Korea (2014)

\bibitem{wainwright2008graphical}
Wainwright, M.J., Jordan, M.I.: Graphical models, exponential families, and
  variational inference. Found. Trends{\textregistered} Mach. Learn.
  \textbf{1}(1--2),  1--305 (Nov 2008)

\bibitem{wang2022yolov7}
Wang, C.Y., Bochkovskiy, A., Liao, H.Y.M.: {YOLOv7}: Trainable bag-of-freebies
  sets new state-of-the-art for real-time object detectors. In: IEEE Conf.
  Comput. Vis. Pattern Recog. pp. 7464--7475. Vancouver, Canada (2023)

\bibitem{honeycombinterobs2012}
Watadani, T., Sakai, F., Johkoh, T., Noma, S., Akira, M., Fujimoto, K.,
  Bankier, A.A., Lee, K.S., M{\"u}ller, N.L., Song, J.W., Park, J.S., Lynch,
  D.A., Hansell, D.M., Remy-Jardin, M., Franquet, T., Sugiyama, Y.:
  Interobserver variability in the {CT} assessment of honeycombing in the
  lungs. Radiology  \textbf{266}(3),  936--944 (Dec 2012)

\bibitem{weilabelsep}
Wei, H., Xie, R., Feng, L., Han, B., An, B.: Deep learning from multiple noisy
  annotators as a union. IEEE Trans. Neural Netw. Learn. Syst. pp. 1--11 (Apr
  2022)

\bibitem{weilabelsepconc}
Wei, J., Zhu, Z., Luo, T., Amid, E., Kumar, A., Liu, Y.: To aggregate or not?
  learning with separate noisy labels. In: ACM SIGKDD Int. Conf. Knowl. Discov.
  Data Min. pp. 2523--2535. Long Beach, CA, USA (2023)

\bibitem{whitehillglad}
Whitehill, J., Wu, T.f., Bergsma, J., Movellan, J., Ruvolo, P.: Whose vote
  should count more: Optimal integration of labels from labelers of unknown
  expertise. In: Adv. Neural Inform. Process. Syst. pp. 2035--2043. Vancouver,
  Canada (2009)

\bibitem{mrprism2021}
Wu, J., Fang, H., Yang, Y., Liu, Y., Gao, J., Duan, L., Yang, W., Xu, Y.:
  Multi-rater prism: Learning self-calibrated medical image segmentation from
  multiple raters (2022), arXiv:2212.00601

\bibitem{yufusionbranch}
Yu, S., Zhou, H.Y., Ma, K., Bian, C., Chu, C., Liu, H., Zheng, Y.:
  Difficulty-aware glaucoma classification with multi-rater consensus modeling.
  In: Med. Imag. Comput. Comput. Assist. Interv. pp. 741--750 (2020)

\bibitem{disgtmedimageseg2020}
Zhang, L., Tanno, R., Xu, M.C., Jin, C., Jacob, J., Ciccarelli, O., Barkhof,
  F., Alexander, D.C.: Disentangling human error from the ground truth in
  segmentation of medical images. In: Adv. Neural Inform. Process. Syst. pp.
  15750--15762 (2020)

\bibitem{zhudataforobjdet}
Zhu, X., Vondrick, C., Ramanan, D., Fowlkes, C.: Do we need more training data
  or better models for object detection? In: Brit. Mach. Vis. Conf. Surrey, UK
  (2012)

\end{thebibliography}
